%% file: hom_dist.tex
\documentclass{article}

\PassOptionsToPackage{sort}{natbib}
\usepackage{iclr2027_conference}
\iclrpreprint
\iclrpreprint
\usepackage{times}
\usepackage{microtype}
\usepackage{graphicx}
\usepackage{booktabs}
\usepackage{wrapfig}
\usepackage{paralist}
\usepackage{algorithm}
\usepackage{algpseudocode}

\usepackage{hyperref}

\usepackage{amsmath}
\usepackage{amssymb}
\usepackage{mathtools}
\usepackage{amsthm}
\usepackage{dsfont}
\usepackage[english]{babel}

\usepackage{paralist}
\usepackage[separate-uncertainty=true,multi-part-units=single]{siunitx}
\usepackage{tikz}
\usetikzlibrary{arrows.meta, calc, decorations.pathreplacing}
\usepackage{standalone}
\usepackage{subcaption}

\usepackage[dvipsnames]{xcolor}

\definecolor{stdblue}{HTML}{D9E0F3}
\definecolor{bestgray}{HTML}{EFEFEF}

\usepackage{enumitem}
\usepackage{tcolorbox}
\usepackage{colortbl}
\usepackage{apptools}
\usepackage{titletoc}

\usepackage{thmtools}
\usepackage{thm-restate}

\definecolor{OIOrange}    {RGB}{230, 159,   0}
\definecolor{OISkyBlue}   {RGB}{ 86, 180, 233}
\definecolor{OIGreen}     {RGB}{  0, 158, 115}
\definecolor{OIYellow}    {RGB}{240, 228,  66}
\definecolor{OIBlue}      {RGB}{  0, 114, 178}
\definecolor{OIVermillion}{RGB}{107,174,214}
\definecolor{OIRose}      {RGB}{33,113,181}

\definecolor{setA}  {RGB}{127,191,123}
\definecolor{setB}  {RGB}{175,141,195}

\definecolor{GLLow}  {RGB}{186,228,179}
\definecolor{GLow}  {RGB}{116,196,118}
\definecolor{GMid}  {RGB}{49,163,84}
\definecolor{GHigh} {RGB}{0,109,44}

\hypersetup{
    colorlinks,
    citecolor = OIVermillion,
    urlcolor  = OIRose,
    linkcolor = OIRose,
}

\declaretheorem[name=Definition, style=definition]{definition}

\usepackage[capitalize, nameinlink]{cleveref}

\usepackage{amsmath,amsfonts,bm}
\usepackage{pict2e}   

\newsavebox{\cmBox}
\newlength{\cmLen}
\newlength{\cmWd}

\DeclareRobustCommand{\colmath}[2]{\begingroup
  \sbox{\cmBox}{$#2$}\setlength{\cmLen}{\dimexpr\wd\cmBox+0.4em-1.1em\relax}\ifdim\cmLen<0pt \setlength{\cmLen}{0pt}\fi
  \setlength{\cmWd}{\dimexpr\cmLen+1.1em\relax}\setlength{\unitlength}{1em}\makebox[0pt][l]{\color{#1}\kern0.55em\relax
    \begin{picture}(0,0)\put(0,0.2){\circle*{1.1}}\end{picture}\rule[-0.35em]{\cmLen}{1.1em}\begin{picture}(0,0)\put(0,0.2){\circle*{1.1}}\end{picture}}\makebox[\cmWd]{\usebox{\cmBox}}\endgroup
  }
\newcommand{\Hom}[2]{\operatorname{Hom}(#1, #2)}
\newcommand{\homcnt}[2]{\operatorname{hom}(#1, #2)}
\DeclareMathOperator{\dis}{dis}
\DeclareMathOperator{\dist}{d}
\DeclareMathOperator{\hd}{d_{HD}}
\DeclareMathOperator{\wl}{d_{WL}}
\DeclareMathOperator{\ddh}{d_{H}}

\DeclareMathOperator{\Ima}{Im}

\newcommand{\dhd}[2]{\hd(#1, #2)}

\newcommand{\dwl}[2]{\wl(#1, #2)}

\newcommand{\hde}[1]{\ensuremath{\Gamma(#1)_{\mathcal{F}}}}

\def\eqref#1{equation~\ref{#1}}

\def\1{\bm{1}}

\DeclareMathAlphabet{\mathsfit}{\encodingdefault}{\sfdefault}{m}{sl}
\SetMathAlphabet{\mathsfit}{bold}{\encodingdefault}{\sfdefault}{bx}{n}

\def\gF{{\mathcal{F}}}
\def\gG{{\mathcal{G}}}

\def\gO{{\mathcal{O}}}

\def\gT{{\mathcal{T}}}

\def\gX{{\mathcal{X}}}

  \title{\emph{Graph Homomorphism Distortion}: A structural (pseudo)-metric for attributed graphs}

\author{Martin Carrasco\thanks{These authors contributed equally.} \\
AIDOS Lab \\
University of Fribourg\\
Fribourg, Switzerland \\
\url{martin.carrascocastaneda@unifr.ch} \\
\And
Olga Zaghen\footnotemark[1] \\
AMLab\\
University of Amsterdam \\
Amsterdam, Netherland \\
\url{o.zaghen@uva.nl} \\
\And
Kavir Sumaraj \\
AIDOS Lab \\
University of Fribourg\\
Fribourg, Switzerland \\
\url{kavir.sumaraj@unifr.ch}
\And
Erik Bekkers \\
AMLab\\
University of Amsterdam \\
Amsterdam, Netherland \\
\url{e.j.bekkers@uva.nl}
\And 
Bastian Rieck \\
AIDOS Lab \\
University of Fribourg\\
Fribourg, Switzerland \\
\url{bastian.grossenbacher@unifr.ch}
}
\begin{document}
\maketitle
\begin{abstract}
	A large driver of the complexity of graph learning is the interplay between \emph{structure} and \emph{features}.
When analyzing the expressivity of graph neural networks, however, existing approaches ignore features in favor of structure, making it nigh-impossible to assess to what extent two graphs with close features should be considered similar.
We address this by developing a new \mbox{(pseudo-)metric} based on graph homomorphisms.
Inspired by concepts from metric geometry, our \emph{graph homomorphism distortion} measures the minimal worst-case distortion that node features of one graph are subjected to when mapping one graph to another.
We demonstrate the utility of our (pseudo-)metric by showing that
\begin{inparaenum}[(i)]
		\item it can be efficiently approximated while retaining distinction capabilities,
		\item it provides a granular approach to graph distinction, and
		\item it  serves as a structural encoding, extending the concept of homomorphism-based descriptors to attributed graphs, thus improving the predictive capabilities of graph neural networks.
	\end{inparaenum}
\end{abstract}

\section{Introduction}

A pivotal concept in graph learning involves assessing the \emph{expressivity} of a model, intended as its ability to distinguish between graphs that are considered to be ``different.''
Typically, ``different'' is taken to mean ``non-isomorphic'' under the Weisfeiler--Leman~(WL) hierarchy of graph isomorphism tests, which remain the \emph{de facto} benchmark for expressivity.
Prior work has linked the WL hierarchy to message-passing neural networks~\citep{xu2018powerful, morris_weisfeiler_2019, morris2023weisfeiler}; however, it is fundamentally limited in that it yields a \emph{binary} quantity and does not account for an underlying measure of similarity, i.e., it cannot capture ``small differences'' between graphs.
This is due to graph isomorphism being a brittle property, meaning that even a single edge modification can break it. 
Moreover, since WL is a \emph{combinatorial} measure, it only captures graph structure while ignoring any features.
The case where features are weighted by importance by using a probability measure has openned the door for more principled considerations, such as using a Gromov-Haussdorf distance as similarity \citep{vayer2020fused}.
As a notable extension of WL to geometric graphs, \citet{joshi2023expressive} develop a model of expressivity based on joint equivalence with respect to graph isomorphism~(\emph{structure}) and Euclidean isometry~(\emph{features}).
While this leads to a principled measure for vertex features with an inherent Euclidean geometry, it still does not account for approximate~(geometric) similarity.
Despite alternatives to strict structural views, like spectral-based methods \citep{isufi2024graph}, being studied, the current ecosystem is still skewed towards structure~\citep{guo2025position} and \emph{should} move away from previous restrictive views of both \citep{vasileiou2026graph}.
\paragraph{Graph homomorphisms: A \emph{melange} of features and structure.}
Our work proposes a more general perspective that compares vertex features while incorporating the graph structure as a restriction.
To this end, we make use of \emph{graph homomorphisms}.
For structural comparisons, graph homomorphism \emph{counts} have been shown to distinguish any pair of non-isomorphic graphs~\citep{lovasz1967operations} while characterizing graph properties~\citep{lovasz_large_2012}.
Thus, homomorphism counts have been successfully used in machine learning~\citep{jin_homomorphism_2024, curticapean_homomorphisms_2017, wolf_structural_2023, bao_homomorphism_2025}, but they still lead to an inherently binary measure of similarity. 
To overcome this, we introduce the concept of \emph{graph homomorphism distortion}, a novel notion of similarity for vertex-attributed graphs inspired by the \emph{Hausdorff distance}, which measures the distance between two subsets of a space via per-point \emph{distortions}.

Through the exploration of our notion of similarity, our paper makes the following key \textbf{contributions}:
\begin{compactitem}
    \item We introduce the graph homomorphism distortion and prove its theoretical properties, including its ability to distinguish between non-isomorphic graphs, induced subgraphs,and strongly symmetrical graphs.
    \item We devise an approximate method to compute our pseudo-metric while retaining strong discriminative capabilities.
    \item We prove that the graph homomorphism distortion \emph{complements} existing expressivity measures like \mbox{$1$-WL} by providing a more granular measure of graph similarity.
    \item We develop a \emph{structural encoding} based on the graph homomorphism distortion and demonstrate its utility in classifying non-isomorphic graphs and serving as a novel \emph{inductive bias} for graph-level tasks.
\end{compactitem}

 \section{Background}\label{subsec:graph_theory}

Our work focuses on  \emph{simple},\footnote{A graph is \emph{simple} if it does not have any self-loops.} \emph{connected}, \emph{undirected} graphs. An undirected graph is tuple denoted  $G = (V, E)$, where  $v \in V$ are the vertices and $\{u,v\} \in E$  the edges.When we deal with multiple graphs at the same time, we denote the vertex set of a graph $G$ with $V(G)$ and the edge set with $E(G)$. The number of vertices (\emph{order}) of a graph is denoted $|G|$. Furthermore, we say that $G' = (V', E')$ is a \emph{subgraph} of $G = (V, E) $, denoted $G' \subseteq G$, if $V' \subseteq V$ and $E' \subseteq E$. If $G'$ contains all edges $\{u', v'\} \in E$ for $u', v' \in V'$ then $G'$ is an \emph{induced subgraph} of $G$, denoted by $G' \coloneq G[V']$.
A \emph{rooted graph} is a tuple $(G,v)$, where $G$ is the graph and $v$ is a fixed node in the graph. The classes of finite graphs with $n$ nodes and finite rooted graphs are denoted as $\mathcal{G}_n$ and $\mathcal{G}_{\bullet}$ respectively.

A \emph{(vertex)-attributed graph}\footnote{For simplicity, we refer to it as an \emph{attributed graph}.}
is a triplet $G = (V, E, f_G)$, where $f_G\colon V \to \gX$ assigns to each vertex $v \in V$ a value $f_G(v)$ in some normed space $(\gX, \|\cdot\|)$.
Throughout this paper, we assume that all attribute functions map into the \emph{same} co-domain. This assumption is commonly satisfied.

For two graphs  $G= (V, E)$ and $H = (V', E')$, a \emph{homomorphism} is a function $\varphi\colon V \to V'$ such that if $\{u,v\} \in E$, then $\{\varphi(u), \varphi(v)\} \in E'$.
A graph $G$ is called a \emph{core} if for every proper subgraph $G'$ there is no homomorphism from $G$ to $G'$. 
We denote the set of all homomorphisms from $G$ to $H$ by $\Hom{G}{H}$ and its cardinality by
$\homcnt{G}{H}$.
The function $\varphi$ is an \emph{isomorphism} if it additionally satisfies that
its inverse  $\varphi^{-1}$ exists and is a homomorphism.
We write $G \simeq H$ to indicate that two graphs are isomorphic. Additionally, two rooted graphs $(G,v)$ and $(H, w)$ are isomorphic if there exists an isomorphism $\varphi$ such that $\varphi(v) = w$. A \emph{graph encoding} is a function $\Gamma\colon \gG \to \mathcal{X}$ that acts on a set of graphs $\gG$ and embeds them into some normed space $\mathcal{X}$. The function $\Gamma$ is an \emph{invariant graph encoding} if it is \emph{permutation-invariant}, i.e., for $G, H \in \gG$ with $G \simeq H$, we have  $\Gamma(G) = \Gamma(H)$. 
Additionally, if for $G \not\simeq H$ we have $\Gamma(G) \neq \Gamma(H)$, we call a graph encoding \emph{complete}.

\section{The Graph Homomorphism Distortion}
\begin{figure}[tbph]
    \centering
    \includestandalone[width=1\linewidth]{figures/intro_fig_tikz}
    \caption{In this example $\mathcal{X} = \mathds{R}^2$, $G=(V, E, f_G)$, $H=(W, E', f_H)$ and $\colmath{setA}{\vphantom{x_1}}$ $\in \mathrm{Im} f_G \subset \mathds{R}^2$ and $\colmath{setB}{\vphantom{x_1}}$ $\in \mathrm{Im} f_H \subset \mathds{R}^2$.
Row $(a)$ shows $\ddh$, the \emph{Hausdorff metric}, evaluated on all pairs, resulting in  $\ddh(G, H) = \max\left(\mathrm{d}_{\gX}(\colmath{setA}{x_3}, \colmath{setB}{y_4}), \mathrm{d}_{\gX}(\colmath{setA}{x_3}, \colmath{setB}{y_3}) \right)$.
Row $(b)$ shows $\mathrm{d_{HD}}$, the \textbf{homomorphism distortion}, where the comparison of a pair is contingent on to the \emph{existence} of a homomorphism. Concretely,  in the $G \to H$ direction, the pair $(\colmath{setA}{x_3},\colmath{setB}{y_4})$ is not accounted for and instead, it is replaced by the pair $(\colmath{setA}{x_3}, \colmath{setB}{y_3})$.
Similarly, in the $H \to G$ direction, the resulting pair is $(\colmath{setA}{x_2}, \colmath{setB}{y_4})$.
In the end, $\ddh(G, H) = \mathrm{d}_{\gX}(\colmath{setA}{x_3}, \colmath{setB}{y_4}) \neq \mathrm{d}_{\gX}(\colmath{setA}{x_2}, \colmath{setB}{y_4}) = \dhd{G}{H}$.
    }
    \label{fig:overview_distortion}
\end{figure} 
The main thread running through this work is on measuring (dis)-similarity between graphs. The only constraint we establish is that the graph theoretic notion of equivalence, isomorphism, shall be respected, i.e., our distance assigns $0$ to two isomorphic graphs. 
Our source of inspiration is the \emph{Hausdorff metric}. For two subsets of a metric space $A,B \subseteq \mathcal{X}$ with metric $\mathrm{d}_{\mathcal{X}}$, it measures the distance between $A$ and $B$ as
\begin{equation}\label{eq:hausdorff}
    \ddh(A,B) \coloneq \max( \sup_{a \in A} \mathop{\vphantom{\sup}\inf}_{b \in B} \mathrm{d}_{\gX} (a,b), \; \sup_{b \in B} \mathop{\vphantom{\sup}\inf}_{a \in A} \mathrm{d}_{\gX} (a,b)).
\end{equation}
Focusing on one of the directions of the interior terms, we interpret it as a \emph{distortion}, where
\begin{equation}\label{eq:pairing}
    \overbrace{\sup_{a \in A} \mathop{\vphantom{\sup}\inf}_{b \in B}}^{\text{pairing}} \underbrace{\mathrm{d}_{\mathcal{X}} (a,b)}_{\text{cost}}.
\end{equation}
This defines a non-symmetric relationship between $A$ and $B$, made symmetric by taking the maximum in each direction.

\paragraph{Homomorphism-restricted distances.}
An attributed graph encodes two kinds of information: features and structure. The Hausdorff metric accounts only for features,  missing the opportunity to exploit structural information.
Our proposal, in contrast, looks at attributed graphs and incorporate the structural information by restricting how we can map one-to-another and \emph{vice versa}. One natural choice of map is a \emph{homomorphism}, which encodes how  a graph can be embedded into another while respecting immediate neighborhoods.
Our \emph{graph homomorphism distortion} aims to encode a notion of distance depending on the vertex attributes of graphs, thus providing a trade-off between expressivity and computation. It consists in a \emph{restriction} of the Hausdorff metric, in terms of which pairs $(a,b)$ for $a \in A$ and $b \in B$ are evaluated, with respect to the existence of a homomorphism. We change \cref{eq:pairing} as follows:
\begin{definition}[Attribute distortion]\label{def:attr_dist}
    Let $G=(V, E, f_G)$ and $H = (V', E', f_H)$ be two attributed graphs with $\mathfrak{G} = \Hom{G}{H}$ the set of all homomorphisms $G \to H$, $f_G \colon V(G) \to \gX$ and $f_H \colon V(H) \to \gX$. Their \emph{attribute distortion} is
    \begin{align}
    \dis (G,H)  &\coloneq \inf_{\varphi \in \mathfrak{G}} \max_{v \in V(G)} \mathrm{d}_{\mathcal{X}}\left( f_G(v), f_H(\varphi(v) \right),
    \end{align} where we set $\inf \emptyset \coloneq \infty$ and $\mathrm{d}_{\gX}
        (x,y) =\| x - y\|$
\end{definition}
\Cref{fig:overview_distortion} illustrates an example. To obtain the symmetric structure of our original muse, the Hausdorff distance, we symmetrize \Cref{def:attr_dist} as follows.
\begin{definition}[Graph homomorphism distortion]\label{def:hom_distortion}
    Given two attributed graphs $G = (V, E, f_G)$ and $H = (V', E', f_H)$, 
    we define their \emph{graph homomorphism distortion} $\hd$ as 
    \begin{equation}\label{eq:d_hd}
        \dhd{G}{H} \coloneq \max\left(  \dis(G,H), \dis(H,G) \right).
    \end{equation}
\end{definition}
In the remainder of the manuscript, we refer to this as \emph{homomorphism distortion}. When we talk about an attributed graph $G$ we implicitly refer to the tuple $(V, E, f_G)$, and we use $\mathfrak{G}$ to denote the set of homomorphisms to another graph $H$.
While \Cref{def:hom_distortion} appears restrictive at first glance, as it requires the homomorphisms to be known explicitly, we will later show how to evaluate it based on families of graphs for which homomorphisms can be efficiently calculated. 

\subsection{General properties of the Homomorphism Distortion}
We first show that the homomorphism distortion is a pseudo-metric~(cf.~\Cref{app:background}) on vertex-attributed graphs. We defer all proofs to  \Cref{app:proofs} for clarity.
\begin{restatable}[Homomorphism distortion is a pseudo-metric]{lemma}{hdpseudometric}
    \label{prop:hdpseudometric}
    The   $\dhd{\cdot}{\cdot}$ is a pseudo-metric, i.e., the following properties hold for all attributed graphs $G$, $G'$, $G''$:
    \begin{compactenum}
        \item $\dhd{G}{G'} \geq 0$ and $\dhd{G}{G} = 0,$
        \item $\dhd{G}{G'} = \dhd{G'}{G},$
        \item $\dhd{G}{G'} \leq \dhd{G}{G''} + \dhd{G''}{G'}.$
    \end{compactenum}
\end{restatable}

The homomorphism distortion is also \emph{$1$-Lipschitz continuous}, i.e., the closer the attribute functions, the smaller the distortion. This property will prove useful for improving the computability of $\hd$.
\begin{restatable}[Lipschitz continuity]{proposition}{lipschitz}\label{prop:lipschitz}
    Given two attribute functions $f,g\colon V \to \gX$ on the same graph $G$, we have 
    $\dhd{(V, E, f)}{(V,E,g)} \leq \|f - g\|_{\infty}$
where $\|\cdot\|_{\infty}$ denotes the supremum norm induced by the norm of $\gX$.
\end{restatable}

\paragraph{Restricting the space of vertex attributes.}
We are interested in attributed graphs whose attribute function is to some extent related to the graph structure. As such, we restrict our focus to functions respecting some desirable properties. Concretely, for an attributed graph $G = (V, E, f_G)$, we expect that if $u,v\in V$ are \emph{automorphic}, then $f_G(u) = f_G(v)$, i.e., $f_G$ respects \emph{global} structure. Similarly, for another graph $H = (W, E', f_H)$,  where $u \in V$ and $w \in W$, if there exists an isomorphism from $G$ to $H$ such that $\varphi(u) = w$, then $f_G(u) = f_H(w)$, i.e., $f_G$ and $f_H$ respect \emph{local} structure.\footnote{The first criterion means that $f_G$ is structurally dependent, whereas the second one means that $f_G$ and $f_H$ are \emph{permutation invariant}.}
To formalize the previous restriction, we need to use the viewpoint of rooted graphs. Recall that the space of all rooted graphs is $\mathcal{G}_{\bullet}$ and their equivalence, isomorphism, is $\simeq_{\bullet}$.
We can enforce both previous restrictions by looking at the quotient space, denoted $\mathcal{G_{\bullet}}/\simeq_{\bullet}$. Consider a function $\Psi$ on the quotient space $\Psi \colon \mathcal{G_{\bullet}}/\simeq_{\bullet} \to  \mathcal{X}$ and let $(G,v), (H,w) \in \mathcal{G}_{\bullet}$ be two rooted graphs. 
It follows immediately that if $G \simeq H$ and $G \neq H$, for any pair of nodes $v \in V(G)$ and $w \in V(H)$ and any isomorphism $\varphi(v) = w$, the rooted graphs will collapse to the same equivalence class. If $G = H$, the claim still holds since an automorphism is also an isomorphism.
The attribute restriction we propose has a name: a \emph{rooted-graph invariant}. This is a function on rooted graphs that evaluates to the same value in case of isomorphism, i.e., $(G,v) \simeq_{\bullet} (H,w) \implies f_G(v) = f_H(w)$.
With this in mind, we introduce one final notational trick in the spirit of rigor and to show how this machinery applies to general graphs:
For a graph $G$, let $\iota_G\colon V(G) \to \mathcal{G}_{\bullet} / \simeq_{\bullet}$ be the quotient map that takes each possible rooted graph in $G$ as $(G,v)$ for $v \in V(G)$ and maps it to its corresponding equivalence class. We define the attribute function on a graph induced by $\Psi$ as 
\begin{equation}\label{eq:f_psi_relation}
    f_G \coloneq \Psi \circ \iota_G, \quad f_G(v) = \Psi\left ([(G, v)] \right).
\end{equation}

Moving forward, we only refer to $\Psi$ to discuss restrictions on the classes
of functions we consider in attributed graphs. Also, we will assume that,
whenever we refer to an attributed graph with a function $f_G$, said function is always
induced by some $\Psi$.
This implies that, if the graph attributes are defined algorithmically, then
they should be obtained through the \emph{same} algorithm on all graphs $G \in
\mathcal{G}_n$. For example, for two attribute functions $f_G,f_H$ on graphs
$G,H$ it cannot happen that $f_G$ is the degree of a vertex and $f_H$ is
constant.
We now turn to illustrate the discriminative power of the homomorphism distortion in terms of isomorphisms.

\subsection{Discriminative power: Non-isomorphic distinction and its flaws }

We circle back to our initial inspiration, \emph{what is the difference between the homomorphism distortion and the Hausdorff metric on the feature space}? Consider a fixed $\Psi$ and the normed space $(\bigcup_{G \in \mathcal{G}}V(G), \mathrm{d}_{\Psi})$ with 
$
\mathrm{d}_{\Psi}(v \in G, w \in H) \coloneqq \mathrm{d}_{\gX}(f_G(v), f_H(w)) = \|f_G(v) - f_H(w)\| = \|\Psi\left ([(G, v)] \right) - \Psi\left ([(H, w)] \right)\|,
$
then the setting described in \cref{eq:hausdorff} is the Hausdorff metric on the normed spaces, i.e.,
\begin{equation}\label{eq:hausedorff_attr}
    \mathrm{d}_{\text{H}}(V(G), V(H)) = \max\left(\sup_{v  \in V(G)} \mathop{\inf\vphantom{\sup}}_{w \in V(H)} \mathrm{d}_{\Psi}(v,w), \sup_{w \in V(H)} \mathop{\inf\vphantom{sup}}_{v \in V(G)} \mathrm{d}_{\Psi}(v,w) \right).
\end{equation}
The following relationship between representation and the homomorphism distortion holds for any scale in terms of the Hausdorff distance.

\begin{restatable}{theorem}{dischausdorff}\label{theo:disc_hausdorff}
    The $\hd$ is \emph{at least as discriminative} as $\Psi$, i.e.,  
    $\ddh(V(G), V(H)) \leq \dhd{G}{H}$.
\end{restatable}
Importantly, $\dhd{G}{H} = 0$ implies that the Hausdorff distance is 0.
The homomorphism distortion is also permutation invariant because of the composition of homomorphisms.
\begin{restatable}[Permutation invariance]{proposition}{perminv}
    \label{prop:perm_inv}
    The $\hd$ is permutation-invariant, i.e., for two attributed graphs $G,H$ with permutations $\rho\colon V(G) \to V(G)$ and $\rho'\colon V(H) \to V(H)$, we have
    $\dhd{\rho(V(G))}{\rho'(V(H))} = \dhd{G}{H}$.
\end{restatable}

Thus, the choice of $\Psi$  provides us with a \emph{knob} that changes its power to distinguish between graphs. This implies that for certain graph encodings, even some simple graphs cannot be distinguished. We illustrate in the following corollaries both ends of the spectrum of expressivity that $\Psi$ provides.
\begin{restatable}{corollary}{psipower}
    \label{coro:psipower}
    The following statements for $\Psi$ hold:
    \begin{compactenum}
        \item If $\Psi$ is constant, then $\dhd{G}{H} = 0\iff \mathrm{core}(G) \simeq \mathrm{core}(H)$.
        \item  If $\Psi$ is injective, then $\dhd{G}{H} = 0\iff G \simeq H$.
        \item If $\Psi$ is injective, then $\hd$ is a metric on the quotient.
    \end{compactenum}
\end{restatable}

Between cases $1.$ and $2.$ there is a range of choices of $\Psi$ which will allow certain graphs to be distinguished and other not. For example, if two graph $G$, $H$ are \emph{co-spectral}, i.e., their adjacency matrices have the same multiset of eigenvalues, then a $\hd$ with $\Psi$ based on the spectrum will not be able to distinguish them.
On the other hand, certain kinds of hard graphs such as the Rook's graph on a $4 \times 4$ grid and the Shrikhande graphs do not share a core, resulting in an infinite distance.
The final scenario, where $\Psi$ is injective and each isomorphism class is assigned a ``canonical'' labeling, is the \emph{graph canonization problem}~\citep{babai1983canonical}. Solving the \emph{graph isomorphism problem} consists in verifying that the output of two ``canonical'' graph representations is equal.

\paragraph{Beyond strict isomorphism.}
Distinction criteria based on isomorphism tests are fundamentally limited when comparing (attributed) graphs: They fail to account for subtle differences that might be defining characteristics in how a label is assigned to a graph. For example, a label could depend on the \emph{core} of a graph and not necessarily on its whole connectivity. This phenomenon is also present in common graph datasets \citep{ivanov2019understanding}. We identify cases where highly similar graphs cannot be distinguished by the homomorphism distortion. In these instances, the high similarity is given by induced subgraphs.
We begin by analyzing the directional distortion between a graph $G$ and an induced subgraph $G'$ of $G$.
\begin{restatable}{proposition}{infdis}
    \label{prop:inf_dis_sub}
    Let $G = (V,E, f_G)$ be an attributed graph and $G' = (V',E', f_{G'})$ be an
    attributed subgraph of $G$. The following statements hold if $f_{G'}$ is a
    restriction of $f_G$ to $V'$:
    \begin{compactenum}
        \item $\dis(G, G') = 0$,
        \item If additionally for any $n\in \mathds{N}$, $G =K_n$ or $G = C_{\ell}$ for odd $\ell > 2 \in \mathds{N}$, then $\dis(G,G') > 0$.
    \end{compactenum}
\end{restatable}  
The distortion \emph{from} an induced subgraph to its supergraph is uninformative, i.e., it is always~$0$.
However, the statement above is asymmetrical in that the distortion \emph{to} an induced subgraph can be nonzero. Note that our overall graph homomorphism distortion $\mathrm{d_{HD}}$ maintains its symmetry, because $\mathrm{d_{HD}}$ is defined as the maximum over both directional infima; cf.\ \cref{eq:d_hd}.
This concludes our analysis of the basic properties of the homomorphism distortion.
In light of recent results from \citet{jin_homomorphism_2024} and \citet{bao_homomorphism_2025}, we investigate the properties and feasibility of an \emph{encoding} consisting of homomorphism distortions to a reference set.

\subsection{Homomorphism Distortion Encoding}

Our main contribution to machine learning is to zoom in on the potential of the homomorphism distortion as a way to \emph{characterize} graphs, or analogously to obtain a graph encoding. 
A robust characterization acts as a fine-grained \emph{signature}, enabling effective comparison between graphs. While permutation invariance is the theoretical ideal condition for distinguishing isomorphic graphs, practical machine learning tasks often require broader capabilities. In this context, encodings have emerged as a standard empirical approach.
Let $\mathcal{F}$ be a family of graphs. 
The homomorphism distortion encoding of an attributed graph $G$ is 
$ \hde{G} = [\dhd{F}{G}]_{F \in \mathcal{F}}$,
where $\hde{G} \in \mathbb{R}^{|\mathcal{F}|}$ is a \emph{signature} of homomorphism distortions from each graph $F \in \mathcal{F}$ to $G$.
We now demonstrate that, to achieve permutation-invariance of our encoding, it is \emph{sufficient} to have $\mathcal{F} = \mathcal{G}_n$.

\begin{restatable}{lemma}{embeddinginvariance}\label{lemma:hom_enc_invariant}
    Let $G \in \mathcal{G}_n$ be an attributed graph. If $\mathcal{F} = \mathcal{G}_n$, then $\hde{G}$ is (permutation) invariant.
\end{restatable}
Notice that \Cref{lemma:hom_enc_invariant} also follows from \cref{prop:perm_inv}. With this, we can extend the previous result to show that the homomorphism distortion encoding is complete.

\begin{restatable}{proposition}{embeddingcomplete}
    \label{prop:hom_enc_complete}
Let $G \in \mathcal{G}_n$ be an attributed graph. If $\mathcal{F} = \mathcal{G}_n$, then $\Gamma(G)_{\mathcal{F}}$ is complete \textit{up to the distinguishing power of $f_G$} \end{restatable}

All the appealing properties of the homomorphism distortion and its promise as an encoding have a drawback: Enumerating homomorphisms, i.e., computing the maps, and computing each of the encoding for a large $\gF$ is intractable.
We remedy this by introducing alternative formulations that make our method computationally feasible.

\subsection{Computational Feasibility}

In practice, computing the number of homomorphisms between two graphs $G, H$ is an NP-complete problem~\citep{boker_complexity_2019}. However, for specific choices or classes of  $G$ and $H$, this problem is considerably easier. 
\citet{daaz_counting_2002} show that counting homomorphisms becomes \emph{computable} in the case of $\Hom{F}{G}$, where $F$ has bounded treewidth $k$. Algorithms that leverage tree decomposition exist~\citep{bodlaender_combinatorial_2007}, thus we rely on dynamic programming solutions to make the computations~\citep{hutchison_fine_2013}.
Nevertheless, we need to ensure that we control the treewidth $k$. We observe that the following relationship hints at a possible strategy to reduce treewidth.
\begin{restatable}{proposition}{subgraphdist}\label{prop:subgraph_dist}
    Let $G,H$ be two attributed graphs and $G',H'$ subgraphs of each one. Then
    \begin{equation*}
        \dhd{G}{H} \geq \dhd{G'}{H'} - \left ( \dhd{G}{G'} + \dhd{H}{H'} \right).
    \end{equation*}
\end{restatable}
If we choose a method for obtaining $G'$, $H'$ that results in a smaller treewidth, we have an effective proxy for the distance.
The utility of \Cref{prop:subgraph_dist} will become apparent when coupled with \Cref{prop:inf_dis_sub} in the experimental section. Here, picking $G',H'$ to be induced subgraphs results in a bound of the form $\dhd{G}{H} \geq \dhd{G'}{H'}$, where we never ``overestimate" the difference in graphs and only underestimate it.\footnote{A consequence is that the distance, as well as the encoding, will remain \emph{invariant} with respect to isomorphism.}
We can also make use of this bound using other reduction methods, like
isospectral reductions \citep{kempton2024isospectral}, where $G',H'$ become
co-spectral and therefore, for an appropriate $\Psi$, also yield $\dhd{G}{G'} =
\dhd{H}{H'} = 0$.

We now turn our attention to the discriminative power of the homomorphism distortion with respect to Weisfeiler--Leman test by framing it as a (pseudo)-metric to allow for comparison.

\section{The WL Distance and Its Shortcomings}

Given that the Weisfeiler--Leman test for graph isomorphism remains the \emph{de facto} benchmark for graph expressivity analysis, we first discuss it and its limitations. We reframe the \mbox{$1$-WL} test in two steps. The first part is a procedure that produces an encoding of two graphs based on their topology. Secondly, it compares their encodings to measure their distance. To begin with, we algebraically formalize what it means for two graphs to be WL-equivalent.

\begin{definition}[Stable \mbox{$1$-WL} coloring encoding]\label{def:stable_coloring}
    A function $c\colon V \to \mathbb{N}$ is a stable \mbox{$1$-WL} coloring encoding if it is the result of applying the \mbox{$1$-WL} algorithm until convergence, i.e., there is no change in colors from one iteration to another. We denote the set of $m$ colors that this procedure yields by $\mathcal{C} = \{1,\dotsc,m\} \subset \mathbb{N}$ and call the multiset
    $h(G) = \{\!\!\{ c(v) | v \in G \}\!\!\}$ the \emph{color histogram} of a graph $G$.
\end{definition}
In the following, we refer to stable $1$-WL colorings as \emph{WL colorings}.
Operating on our previous definitions of vertex-attribution, during this
section, we fix $\Psi$ to be a WL coloring and let the resulting attribute
functions, i.e., $f_G$, reflect said coloring such that for a graph $G$, $\Ima f_G = h(G)$.
It then follows that the homomorphism distortion between graphs with a WL coloring can distinguish \emph{at least as many} non-isomorphic graphs as the WL test.

\begin{restatable}[Generalization of WL distinguishability]{proposition}{hdwl}
    Let $G$ and $H$ be two non-isomorphic graphs whose attribute functions $f_G=c$ and $f_H=c'$ correspond to their respective WL colorings. If WL can distinguish them, then $\dhd{G}{H} > 0$.
\end{restatable}

In the following, we show that if we define a \emph{(pseudo)-metric} associated to the two tests, a similar relation holds between them.
To start with, we consider the equality comparison of two color histograms, on which WL equivalence is based.

\begin{definition}[WL (pseudo)-metric]\label{def:hist}
    Let $G,H$ be two graphs with $\Psi$ a WL coloring and $h(G), h(H)$ the corresponding color histograms.
    This leads to a discrete (pseudo)-metric
    \begin{equation}\label{eq:d_wl}
        \dwl{G}{H} = \begin{cases}
            0 & \text{if } h(G) = h(H), \\
            1 & \text{otherwise.}
        \end{cases}
    \end{equation}
\end{definition}

This distance is too coarse for common machine-learning applications. We thus extend it to use the WL encoding under a metric that compares multisets.
Following \cref{def:hist} and \cref{theo:disc_hausdorff} we now establish an \emph{ordering} of these metrics, which we interpret as their discriminative power.

\begin{restatable}[Discriminative power of coloring]{theorem}{discpowercolor}\label{theo:discpower2}
    Let $G$ and $H$ be two graphs and $\Psi$ be a WL coloring. Then 
    $
        \dwl{G}{H} \leq \ddh(V(G), V(H)) \leq \dhd{G}{H}
    $.
\end{restatable}

Thus, the $\wl$ (pseudo)-metric is the \emph{least} discriminative due to the information loss inherent in binary comparisons. While our extension in terms of the Hausdorff metric, $\mathrm{d}_{\text{H}}$, provides a richer signal in the form of a \emph{continuous} measurement, $\hd$ achieves the highest discriminative power: It simultaneously accounts for both the structural attributes and the homomorphisms between $G$ and $H$.
This ordering of discriminative power implies a relationship between the induced topological spaces. Specifically, we obtain an inclusion of the associated topologies:
\begin{restatable}[Granularity of metric topologies]{proposition}{metrictopol}
    Let $\mathcal{T}_{\mathrm{HD}}$ and $\mathcal{T}_{\mathrm{WL}}$ be the metric topologies induced by $\hd$ and $\mathrm{d}_{\mathrm{WL}}$, respectively. Then, on a set of attributed graphs $\mathcal{G}$ with $\Psi$ as the WL coloring, we have $\mathcal{T}_{\mathrm{WL}} \subset \mathcal{T}_{\mathrm{HD}}$.
\end{restatable}

Intuitively, this inclusion holds because a ``stronger'' pseudo-metric like $\hd$, which assigns larger distances, generates smaller, more selective open balls. The $\mathcal{T}_{\mathrm{HD}}$ topology is thus \emph{finer}, meaning it distinguishes graph structures more precisely than $\mathcal{T}_{\mathrm{WL}}$. In terms of convergence, this implies that the condition for $\hd$ is stricter: Any sequence of graphs converging in the HD topology must necessarily converge in the WL topology, but not vice versa.

 \section{Experiments}\label{sec:results}

Our theoretical analysis highlights the utility of the homomorphism distortion both as a \emph{notion of difference} and as a \emph{graph encoding}. 
In this section, we empirically validate these properties.\footnote{Please refer to \cref{app:extended_experiments} for hardware details and the full list of hyperparameters.}
Specifically, we aim to answer the following research questions:
\begin{enumerate}[noitemsep,left=0.5pt, topsep=0pt,parsep=0pt,partopsep=0pt,label=Q\arabic*]
    \item  \label{q:1} Can the homomorphism distortion  distinguish non-isomorphic highly symmetric graphs ?
\item How do the features in the attributed graphs affect the homomorphism distortion? \label{q:2}
    \item Does the homomorphism distortion encoding serve as a useful inductive bias? \label{q:3}
    \item Does approximation by treewidth decomposition work in practice? \label{q:4}
\end{enumerate}

\paragraph{Functions $\Psi$ over the graphs.}

The theoretical results require the attribute function $\Psi$ to be a rooted-graph invariant. Thus, we use known \emph{positional and structural encodings}~\citep[PSEs]{kanatsoulis2025learning}, specifically,
\begin{inparaenum}[(i)]
    \item Random Walk positional encodings~\citep[RW]{dwivedi2021graph}, which use probabilities of random walks to characterize local structural roles,
    \item Shortest Path encodings~(SP), which use sorted vectors of shortest-path distances to other vertices, and
    \item Effective Resistance encodings~(ER), which assign a sorted vector of resistances from each vertex to each other.
\end{inparaenum}

\paragraph{Computing distortion between large graphs.}

By \Cref{prop:subgraph_dist}, we can approximate the true distance by constructing subgraphs that have a smaller treewidth and are thus computable. We restrict ourselves to $\mathrm{tw}(G) \in \{1, 2\}$ due to computability reasons and ensure all graphs satisfy in either case.
For $\mathrm{tw}(G)=1$ we take the \emph{minimum spanning tree} of $G$, i.e., the smallest tree that ensures $G$ is connected, and for $\mathrm{tw}(G) = 2$, we start with the vertex set $V$ and progressively add edges from $E$ that do not increase the treewidth past the threshold. If all of the remaining edges increase the treewidth, we terminate the process.
The computational complexity of calculating $\mathrm{d_{HD}}$ is $\gO(m^{k+1}\cdot k^2 \log k \cdot n)$, where $k \coloneq \mathrm{tw}(G)$ and $m \coloneq |H|$, for a target graph $H$. See \Cref{app:extended_experiments} for a breakdown.

\subsection{Non-learning Scenarios}
\setlength{\intextsep}{0.05pt}
\captionsetup[table]{skip=2pt, position=top}
\begin{wraptable}[14]{r}{0.45\textwidth}
    \centering
    \caption{The \% of distinguished graphs in
        \texttt{BREC}.}
    \label{tbl:brec}

    \sisetup{
        detect-weight=true,
        table-comparator=true,
        table-format=3.1,
        mode=text
    }

    \setlength{\tabcolsep}{2.5pt}
    \renewcommand{\arraystretch}{0.90}

    \resizebox{\linewidth}{!}{\begin{tabular}{
            @{\hspace{4pt}}
            l
            S
            S
            S
            S
            S
            S
            @{\hspace{4pt}}
        }
            \toprule
            Method
            & {\text{B}}
            & {\text{R}}
            & {\text{E}}
            & {\text{C}}
            & {\text{4}}
            & {\text{D}} \\
            \midrule
            1-WL
            & 0.0
            & 0.0
            & 0.0
            & 0.0
            & 0.0
            & 0.0 \\
            3-WL
            & \bfseries 100.0
            & 37.0
            & \bfseries 100.0
            & 60.0
            & 0.0
            & 0.0 \\
            PH
            & \bfseries 100.0
            & \bfseries 97.0
            & \bfseries 100.0
            & 6.0
            & \bfseries 100.0
            &  5.0 \\
            \midrule
            $\Psi$ & \multicolumn{6}{c}{$\mathrm{tw}(G) = 1$} \\
            \cmidrule{1-7}
            SP
            & 26.7
            & 13.0
            & 41.0
            & 12.0
            & 0.0
            & 0.0 \\

            RW
            & \bfseries 100.0
            & 55.0
            & \bfseries 100.0
            & 13.0
            & 0.0
            & \bfseries 15.0 \\
            ER
            & \bfseries 100.0
            & 50.0
            & \bfseries 100.0
            & \bfseries 94.0
            & 0.0
            & 0.0 \\
            \\
            $\Psi$ & \multicolumn{6}{c}{$\mathrm{tw}(G) \leq 2$} \\
            \cmidrule{1-7}
            SP
            & 58.3
            & 14.0
            & 51.0
            & 12.0
            & 0.0
            & 0.0 \\

            RW
            & \bfseries 100.0
            & 57.0
            & \bfseries 100.0
            & 13.0
            & 0.0
            & \bfseries 15.0 \\
            
            ER
            & \bfseries 100.0
            & 50.0
            & \bfseries 100.0
            & \bfseries 94.0
            & 0.0
            & 0.0 \\
            \bottomrule
        \end{tabular}}
\end{wraptable}
 
We measure the discriminative power, the impact of features in attributed graphs and the effectiveness of treewidth decomposition (\ref{q:1}, \ref{q:2}, \ref{q:4}) by employing the expressivity benchmark dataset \texttt{BREC}~\citep{wang_empirical_2024}. 
The task on this dataset is to distinguishing between highly similar non-isomorphic graphs. Following the methodology of \citet{ballester_expressivity_2024}, we establish a threshold under which graphs are isomorphic as $\texttt{float32}$ machine precision, i.e., $\epsilon = $\num{1.19e-7}.
\Cref{tbl:brec} shows the results in distinguishing each class of graphs. 

\paragraph{The $\hd$ distinguishes non-isomorphic graphs.} 
Our method performs particularly well on CFI $(\mathrm{C})$ and distance-regular $(\mathrm{D})$ graphs compared to non-GNN method $3$-WL~\citep{wang_empirical_2024} and results from \citet{ballester_expressivity_2024}, obtained using persistent homology~(PH). Naturally, $1$-WL cannot distinguish any graphs. 
Notice that PH outperforms all other methods in the case of regular and strongly-regular $(\mathrm{R})$ and 4-distance regular $(\mathrm{4})$.
These results leads to a strong positive answer to \ref{q:1}~$\checkmark$, with ER, followed by RW as the most effective encodings. 
This phenomenon provides insight into \ref{q:2}, indicating the role that the choice of $\Psi$ plays in the performance of the homomorphism distortion.

\paragraph{Decompositions are useful.}
For graphs with a treewidth larger than $3$, enumerating homomorphisms is \emph{intractable}. Thus, we have to rely on approximations. The quality of the approximation and the speedup it provides are inevitably related. We observe that in our experiments the trade-off is minimal. For $\mathrm{tw}(G) = 2$, computations take $3$ hours; for $\mathrm{tw}(G)=1$, computations only take $6$ minutes, which is $\times 28$ faster, with a big and slight degradation on $\mathrm{SP}$ and $\mathrm{RW}$, respectively.
Overall, both decompositions provides reliable differentiation in addition to making the computations tractable with provides positive evidence for \ref{q:4}~$\checkmark$.

\subsection{Learning Scenarios}
We address the remaining research questions~(\ref{q:2}, \ref{q:3}, \ref{q:4}) by using the homomorphism distortion encoding in a set of machine learning tasks.
We focus on graph level tasks, since we propose a graph-level encoding, 
We add the signature information resulting from $\Gamma(G)$ as graph-level information. We embed this signature in a high-dimensional space using a function $\omega \colon \mathds{R}^M \to \mathds{R}^d$, which we parametrize with a 2-layer MLP.

\setlength{\intextsep}{5pt}
\captionsetup[table]{skip=5pt, position=top}

\begin{table}[bp]
\caption{The MAE~($\downarrow$) for graph regression on \texttt{ZINC-12k}~(without edge features) using $\mathcal{F} = \{C_i \mid 3 \leq i \leq 8\}$ for comparability. Hom, Sub and Spasm are from \citet{jin_homomorphism_2024} and $\mathrm{MoSe}$ from \citet{bao_homomorphism_2025}.
Super-indices $\dagger$ and $\ast$ denote the addition of homomorphism and subgraph counts as node features, respectively. Standard deviations are given in the last digit, e.g., $0.380(9) = 0.380 \pm 0.009$.}
\label{tbl:zinc_cycle_no_edge_feat}
\sisetup{detect-all=true,detect-weight=true,table-align-uncertainty=true,
         uncertainty-mode=compact,table-format=1.3(2),tight-spacing=true}
\centering
\resizebox{\linewidth}{!}{\begin{tabular}{@{}l S S S S S S S S S@{}}
    \toprule
     & & \multicolumn{5}{c}{\emph{Count-based}} & \multicolumn{3}{c}{\emph{Ours}} \\
    \cmidrule(lr){3-7} \cmidrule(l){8-10}
     & {Base} & {$\mathrm{Hom}^\dagger$} & {$\mathrm{Sub}^\ast$}
     & {$\mathrm{Spasm}^{\circledast}$} & {Hom+F} & {MoSe}
     & {$\Gamma(G)$} & {$\Gamma(G)^\dagger$} & {$\Gamma(G)^{\ast}$} \\
    \midrule
    GAT & 0.380(0.009) & 0.229(0.008) & 0.201(0.004) & 0.147(0.004) & {--}         & {--}         & 0.317(0.012) & 0.183(0.005) & 0.168(0.003) \\
    GCN & 0.282(0.007) & 0.235(0.005) & 0.198(0.003) & 0.165(0.004) & 0.207(0.008) & {--}         & 0.231(0.016) & 0.189(0.006) & 0.164(0.003) \\
    GIN & 0.246(0.019) & 0.197(0.015) & 0.143(0.002) & 0.143(0.004) & 0.174(0.005) & 0.158(0.004) & 0.214(0.016) & 0.174(0.006) & 0.131(0.003) \\
    \bottomrule
    \end{tabular}}
\end{table} 
\paragraph{Evaluation.}
We evaluate molecular datasets \texttt{ZINC-12K}, with the same setup as \citet{jin_homomorphism_2024},
which follows \citet{dwivedi2023benchmarking},  \texttt{ogbg-molhiv} following \citet{hu2020open} and \texttt{NCI1}  following \citet{morris2020tudataset}. We choose these due to optimal treewidth properties of the graphs in them~(cf.\ \Cref{tbl:treewidths_ds}).
For the remaining ones, we approximate the distances using the method in the previous subsection.
We choose three GNN models~(GAT, GCN, and GIN) and \num{5} random seed initializations, reporting the mean and standard deviation~($\mu \pm \sigma$). We choose the best hyperparameters based on the validation performance and report the scores on the held-out test set.
Our comparison partners are HC~\citep{wolf_structural_2023}, Num~\citep{nguyen2020graph} and $\mathrm{Hom}+F$~\citep{welke2023expectation}.

\setlength{\intextsep}{0.1pt}
\begin{wraptable}[15]{r}{0.48\textwidth}
\caption{
Performance of homomorphism-based encodings with GIN.
}
\label{tbl:nci1_comp}
\centering
\sisetup{detect-all=true,detect-weight=true,table-align-uncertainty=true,table-comparator=true,table-format=<2.2(2.1)}
\begin{small}
\begin{tabular}{@{\hspace{5pt}} l S S @{}}
    \toprule
    Method & {\texttt{NCI1}} & {\texttt{ogbg-molhiv}} \\
           & {ACC ($\uparrow$)} & {AUROC ($\uparrow$)} \\
    \midrule
    $\mathrm{Base}$ & 80.73(1.87) & 77.00(1.49) \\
    \midrule
    \multicolumn{3}{@{}l}{\textit{Count-based}} \\
    $\mathrm{HC}$    & 77.30(4.00) & 74.70(1.00) \\
    $\mathrm{Num}$   & 76.67(1.16) & {--}        \\
    $\mathrm{Hom}+F$ & {--}        & 77.70(1.50) \\
    \midrule
    \multicolumn{3}{@{}l}{\textit{Ours}} \\
$\Gamma(G)_{\mathrm{RW}}$ & 81.95(1.30) & 77.35(1.59) \\
$\Gamma(G)_{\mathrm{ER}}$ & 81.68(1.40) & 77.10(1.11) \\
$\Gamma(G)_{\mathrm{SP}}$ & 82.46(1.26) & 77.85(1.06) \\
\bottomrule
\end{tabular}
\end{small}
\end{wraptable} 
\paragraph{The $\hd$ is an effective inductive bias and it complements homomorphism counts.} \Cref{tbl:zinc_cycle_no_edge_feat} presents the results on \texttt{ZINC-12K} against results the state-of-the-art methods using homomorphism-based inductive biases.
Using the homomorphism distortion encoding yields in better performance than the baseline and it is  competitive with plain homomorphism counts across \emph{all} models. Moreover, complementing our method with either $\mathrm{Hom}$ or $\mathrm{Sub}$ counts yields the best performance in GIN.
Our interpretation is that count and homomorphism distortion information, while competitive among each other, show to be \emph{complementary} as an inductive bias. 
\Cref{tbl:nci1_comp} presents the results on additional molecular datasets where the homomorphism distortion with $C_9$ outperforms all plain count-based methods.
We interpret these results as positive confirmation of~\ref{q:3}~$\checkmark$; additionally they provide further evidence for~\ref{q:4}~$\checkmark$. \section{Discussion}\label{sec:conclusions}

Our paper questions the current expressivity paradigm in graph representation learning and highlights some of its shortcomings, such as the reliance on purely structural comparisons:
As a first step towards addressing these limitations, we introduce the graph homomorphism distortion, a novel measure of dissimilarity for attributed graphs.
This (pseudo-)metric captures a continuous notion which fully characterizes graphs under certain assumptions, 
inducing a metric on the space of isomorphism equivalence classes.
Moreover, we show that it complements existing expressivity measures such as \mbox{$1$-WL} in terms of their topological granularity and discriminative power.
By reframing the homomorphism distortion as a \emph{structural encoding}, we obtain an encoding that
\begin{inparaenum}[(i)]
    \item fully characterizes graphs in \emph{theory}, 
    \item is computable, and 
    \item improves predictive performance,  alone or as an addition to homomorphism counts, when used as an inductive bias in \emph{practice}.
\end{inparaenum}

\paragraph{Limitations and future work.} Our formulation suffers from some limitations, which we aim to address in future work.
Most notably, our measure currently requires general assumptions on the vertex-attributed graphs to guarantee completeness, and obtaining more specific results would be desirable.
Moreover, for practical reasons, we refrain from using more structural or positional encodings as vertex attribute functions; we believe that more suitable functions exist or may be learned.
A thread along this paper is developing a computational method for analyzing graph expressivity, akin to WL, we leave this for future work.
We also postpone additional theoretical inquiries concerning which graph properties~(e.g., the presence/absence of cycles) are captured by different proxy graph families~$\mathcal{F}$, and instead restrict our experiments to a limited exploration of such classes.
We believe that a \emph{substantial shift} in graph representation learning is required to move beyond restricted expressivity measurements.
While the community is already~(partially) aware of these flaws~\citep{bechlerspeicher25a, coupette25a}, additional steps need to be taken to fix them.
Our work proposes one particular direction in the search for better expressivity measurements. 
\clearpage

\subsection*{AI use statement}

In this work, we used generative AI tools for obtaining prototype code for submitting tasks to our HPC infrastructure. We also used generative AI tools to create TikZ figure source code and formatting tables. We have not used AI tools for any other tasks with required or recommended disclosure.
We have checked and refined all AI-generated outputs and take full responsibility for the contents of this work.

\subsection*{Reproducibility statement}

In Section 3.4 we present a method for tackling computational feasibility. In Section 5.1 we present the method for obtaining lower treewidth graphs, denoted \textbf{Computing distortion between large graphs}. In Section 5.1 we present results on \texttt{BREC} and along that the choices of threshold, $\epsilon$, and the chosen treewwidths. In Section 5.2 we reference the method to encode the distortion vector and what it consists of. In Section B and C we provide proofs for all of our propositions, lemmas and theorems. In Section D we reference: hardware details used for the computations in \textbf{Hardware setup}, hyperparameters we sweep through in \textbf{Hyperparameters}, and a pseudo-code sketch of the computation of the homomorphism distortion in \textbf{Algorithm computations}.

\bibliography{biblio}
\bibliographystyle{iclr2027_conference}

\onecolumn
\appendix
\crefalias{section}{appendix}

\makeatletter
\def\addcontentsline#1#2#3{\addtocontents{#1}{\protect\contentsline{#2}{#3}{\thepage}{\@currentHref}}}
\makeatother

\startcontents
\printcontents{}{1}{{\hrule
			\large\textbf{Appendix~(Supplementary Materials)}\vskip3pt\hrule\vskip5pt}
}
\vskip10pt
\hrule
\vskip5pt

\setcounter{figure}{0}
\setcounter{table}{0}
\renewcommand{\thefigure}{S.\arabic{figure}}
\renewcommand{\thetable}{S.\arabic{table}}

\section{Extended Background}\label{app:background}

We subsequently provide more information about the mathematical preliminaries and graph theory, which serve to make this paper self-contained.

\subsection{Mathematical Preliminaries}
\begin{definition}[Pseudo-metric]\label{def:pseudometric}
    A pseudo-metric is a tuple $(X, \dist)$ for a set $X$ and a function $\dist\colon X \times X \to \mathds{R}_{\geq 0}$ where for all $x,y,z \in X$ the following hold:
    \begin{compactenum}
        \item $\dist(x, x) = 0$
        \item $\dist(x, y) = \dist(y, x)$
        \item $\dist(x, z) \leq \dist(x, y) + \dist(y, z)$
    \end{compactenum}
\end{definition}

\begin{definition}[Metric]\label{def:metric}
    A metric is a tuple $(X, \dist)$ for a set $X$ and a function $\dist\colon X \times X \to \mathds{R}_{\geq 0}$ where for all $x,y,z \in X$ the following hold:
    \begin{compactenum}
        \item $\dist(x, x) = 0$
        \item $x \neq y \Rightarrow \dist(x, y) > 0$
        \item $\dist(x, y) = \dist(y, x)$
        \item $\dist(x, z) \leq \dist(x, y) + \dist(y, z)$
    \end{compactenum}
\end{definition}
From the previous definitions a metric is always a pseudo-metric. Additionally, we call a a pseudo-metric \emph{extended} if it can take values in $[0, \infty]$, i.e. additing infinity.

\begin{definition}[Metric Topology]\label{def:metric_topology_appendix}
    Let $(X, d)$ be a metric space. The open ball of radius $\epsilon > 0$ centered at $x \in X$ is defined as:
    \begin{equation}
        B_d(x, \epsilon) = \{ y \in X \mid d(x, y) < \epsilon \}.
    \end{equation}
    The \textit{metric topology} induced by $d$, denoted $\mathcal{T}_d$, is the topology generated by the basis of all open balls. A set $U \subseteq X$ is open in $\mathcal{T}_d$ if and only if for every $x \in U$, there exists an $\epsilon > 0$ such that $B_d(x, \epsilon) \subseteq U$.
\end{definition}

\subsection{Graph Theory}

We briefly expand on several nonstandard concepts from graph theory, namely \emph{graph cores} and \emph{treewidth}, both of which will be crucial components in the subsequent proofs.

\begin{definition}[Tree decomposition]\label{def:treewidth}
    A \emph{tree decomposition} of a graph $G$ is a pair $(\gT, \{X_a\}_{a \in V(\gT)})$ where $\gT$ is a tree and $\{X_a\}_{a \in V(\gT)}$ is the family of subsets---referred to as bags---of $V(G)$ where the following conditions hold:
\begin{compactenum}
        
        \item Each $v \in V(G)$ is in at least one bag $X_a$.
        \item For each pair $\{v, u\} \in E(G)$, there is at least one bag $X_a$ such that $v, u \in X_a$.
        \item For each $v \in V(G)$ the set $\gT_v := \{a \in V(\gT) | v \in X_a\}$ induces a connected subgraph of $\gT$.
    \end{compactenum}
\end{definition}

The width of a tree decomposition is $\max_{a \in V(\gT)} |X_a| - 1$, the maximum size of a \emph{bag} minus $1$. The minimum width among all possible decompositions is called the \emph{treewidth} of $G$, denoted $\mathrm{tw}(G)$.

 \section{Proofs of Homomorphism Distortion Properties}\label{app:proofs}

\hdpseudometric*
\begin{proof}
    Since norms are non-negative, we have that $\dhd{G}{G'} \geq 0$. If $G=G'$, the identity mapping $f(v) = v $ is in $\Hom{G}{G}$, minimizing the right-hand side of \Cref{def:hom_distortion}, which evaluates to $0$. Thus, Item 1 holds. As for the symmetry, \Cref{def:hom_distortion} evaluates terms involving $\Hom{G}{ G'}$ and $\Hom{G'}{G}$ respectively, and taking their maximum. This operation is inherently symmetric, so switching $G$ and $G'$ has no effect. Thus, $\hd$ is symmetric and Item 2 holds. Finally, to prove the triangle inequality, i.e., Item 3, we make use of the fact that we can compose homomorphisms. Without loss of generality, we assume that $G, G'$, and $G''$ are homomorphically equivalent, i.e., there is at least one homomorphism between arbitrary pairs of these graphs. Taking any two such homomorphisms $\varphi\colon G \to G''$ and $\varphi'\colon G'' \to G'$  their composition $\varphi' \circ \varphi$  is a homomorphism from $G$ to $G'$. As such, it occurs in the calculation of $\dhd{G}{G'}$. Due to the properties of the norm used in \Cref{def:hom_distortion}, splitting the composition into $\varphi$ and $\varphi'$ is always lower bounded by taking the infimum over all homomorphisms in $\dhd{G}{G'}$. Hence, the triangle inequality described in Item 3 holds.
\end{proof}

\lipschitz*
\begin{proof}
    Since we are dealing with the same graph, the identity is in $\Hom{G}{G}$. If we evaluate \Cref{def:hom_distortion} with the identity, we have that $\max_{v \in V} \| f(v) - g(v)\|$ appears in the right-hand side minimization. Since the infimum is taken for calculating \Cref{def:hom_distortion} the claim follows. 
\end{proof}

\dischausdorff*
\begin{proof}

Let $\Psi$ be a function on $\mathcal{G}_{\bullet}$ to some normed space $\mathcal{X}$ as defined previously. For two attributed graphs $G,H$ this is exactly
\begin{equation}
    \| f_G(v) - f_H(w)\| = \|\Psi\left ([(G, v)] \right) - \Psi\left ([(H, w)] \right)\|,
\end{equation}

then, the $\mathrm{d}_{\text{H}}(V(G),V(H))$ is given by

$$
\ddh(V(G), V(H)) = \max \left( \sup_{v \in V(G)} \mathop{\vphantom{\sup}\inf}_{w \in V(H)} \mathrm{d}_{\gX}(v,w), \sup_{w \in V(H)} \mathop{\vphantom{\sup}\inf}_{v \in V(G)} \mathrm{d}_{\gX}(v,w) \right)
$$
and the one-way distortion is given by 

$$
\mathrm{dis}(G,H) = \mathop{\vphantom{\sup}\inf}_{\varphi \in \mathfrak{G}} \max_{v \in V(G)} \|f_G(v) - f_H(\varphi(v)) \|.
$$
We focus our attention on 
$$
    \sup_{v \in V(G)} \mathop{\vphantom{\sup}\inf}_{w \in V(H)} \| f_G(v) - f_H(w)\|.
$$
First, we clarify some notation. A relationship $R \subseteq A \times B$ is a set containing some number of pairigs of elements of $A$ and $B$. For a relationship $R \subseteq A \times B$ and a fixed element $a \in A$, we denote all pairs with $a$ as the first element with $R(a, B) = \{ (a,b) : b \in B\} \subseteq R$. Alternatively, we can see it as the set containing pairings of $a$ with any $b \in B$. The Hausdorff distance operates on the following relationship
$$
Q \coloneq V(G) \times V(H),
$$ i.e all possible pairings $(u,v)$ for $v \in V(G), w \in V(H)$. In contrast, assume that $G$ and $H$ are homomorphic and let $\varphi \in \Hom{G}{H}$ be any homomorphism $\varphi \colon V(G) \to V(H)$. The pairing relationship for $\varphi$ is
$$
R_{\varphi} \coloneq  \{(v, \varphi(v)) : u \in V(G) \}.
$$
Naturally, $R_{\varphi} \subseteq Q$ for any $\varphi$ since we are restricting ourselves to less pairings, i.e., the ones that fulfill the homomorphism criteria. Now fix any $v \in V(G)$, the contention of the pairings holds
$$
R_{\varphi}(v, V(H)) \subseteq Q(v, V(H)).
$$
The infimum can then be bounded in terms of these subsets in terms of $v$ along with all $w \in V(H)$ as
\begin{align*}
    \inf_{w \in V(H)} \| f_G(v) - f_H(w) \| &= \inf_{(v,w) \in Q(v, V(H))} \| f_G(v) - f_H(w) \| \\
    & \leq \inf_{(v,\varphi(v)) \in R_{\varphi}(v, V(H))} \| f_G(v) - f_H(w)\| \\
    &= \| f_G(v) - f_H(\varphi(v)) \|,
\end{align*}
where we arrive to the final equality since $R_\varphi(v, V(H))$ is the set with one element, namely $\{(v, \varphi(v))\}$. Since we chose $v$ arbitrarily this holds for any $v \in V(G)$. Now, we can evaluate the outer supremum noting
$$
    \| f_G(v) - f_H(\varphi(v)) \| \leq \sup_{u \in V(G)} \| f_G(u) - f_H(\varphi(u)) \|
$$
thus for $v \in V(G)$ it holds that 
\begin{align*}
    \inf_{w \in V(H)} \| f_G(v) - f_H(w)\| \leq \| f_G(v) - f_H(\varphi(v)) \| \leq \sup_{u \in V(G)} \| f_G(u) - f_H(\varphi(u)) \| 
\end{align*}
since this holds for all $v \in V(G)$, then it also holds for the supremum
\begin{align*}  
    \sup_{u \in V(G)} \mathop{\vphantom{\sup}\inf}_{w \in V(H)} \| f_G(u) - f_H(w)\| &\leq \sup_{u \in V(G)} \| f_G(u) - f_H(\varphi(u)) \| = \max_{v \in V(G)} \| f_G(u) - f_H(\varphi(u)) \| 
\end{align*}
Since $V(G)$ is finite we have equality between $\sup \text{ and } \max$.  Furthermore the above equation  holds for  any $\varphi \in \Hom{G}{H}$ thus we may conclude 
\begin{align*}
     \sup_{u \in V(G)} \mathop{\vphantom{\sup}\inf}_{w \in V(H)} \| f_G(u) - f_H(w)\| &\leq \mathop{\vphantom{\sup}\inf} _{\varphi \in \mathfrak{G}}\max_{v \in V(G)} \| f_G(u) - f_H(\varphi(u)) \|.
\end{align*}

If $\Hom{G}{H} = \emptyset$ then we get $\infty$ and the bound still holds. We can provide a similar argument for the other direction and taking the max over both
preserves the maximum relationship. The claim follows from this final fact.
\end{proof}

\perminv*

\begin{proof}
Let $G=(V,E,f_G)$ and $H=(V',E',f_H)$. For arbitrary permutations $\rho\in \mathrm{S}_{|V|}$ and $\gamma\in \mathrm{S}_{|V'|}$,  let $G' = (\rho(V), E_{G'},f_{G'})$ and $H' = (\gamma(V'), E_{H'},f_{H'})$. Since $f_{G'}=\Psi\circ\iota_{G'}$ and $\iota_{G'}=\iota_G\circ \rho^{-1}=\iota_G$  we have $f_{G'}=\Psi\circ\iota_{G'} =\Psi\circ\iota_{G}= f_G$. Likewise we have $f_H = f_{H'}.$

The application of a permutation on a graph results in an isomorphic graph thus, if $G \simeq H$ then $\dhd{G}{H} = \dhd{G'}{H'} = 0$. Suppose $G \not\simeq H$ and $\varphi\colon G \to H$ is any homomorphism. Note $\rho^{-1}:G'\rightarrow G$ and  $\gamma^{-1}:H'\rightarrow H$ thus we may define $\psi = \gamma\circ \varphi \circ \rho^{-1} \in \Hom{G'}{H'}$.  

For any  vertex $w \in \rho(V)$ there exists a vertex $v\in V$ such that $f_{G'}(w) = f_{G'}(\rho(v)) = f_G(v)$ since $f_{G'}$ is permutation invariant. Additionally $f_{H'}(\psi(w)) = f_{H'}(\gamma\circ\varphi\circ\rho^{-1}(\rho(v))) =f_H(\varphi(v))$. Combining these we obtain   
    \begin{align*}
         \max_{w\in \rho(V)} \| f_{G'}(w) - f_{H'}(\psi(w)) \| 
        &= \max_{v \in V} \| f_G(\rho(v)) - f_{H}(\gamma \circ \varphi (v)\| = \max_{v \in V} \| f_G(v) - f_H(\varphi  (v))\|  .
    \end{align*}
Let $\mathfrak{G}= \Hom{G}{H}$ and $\mathfrak{G}'=\Hom{G'}{H'}$. Recall 
\begin{align*}
     \dis(G,H)= \inf_{\varphi\in \mathfrak{G}}  \max_{v \in V} \| f_G(v) - f_H(\varphi(v)) \|.
\end{align*} Assume for the fixed graph homomorphism $\varphi'\in \mathfrak{G}$, $\varphi'$ defines the distortion i.e.,
\begin{align*}
     \dis(G,H)=    \max_{v \in V} \| f_G(v) - f_H(\varphi'(v)) \|.
\end{align*}
If we consider $\psi'= \gamma\circ\varphi'\circ\rho^{-1} \in \mathfrak{G}'$ then 
\begin{align*}
    \dis(G,H) &= \max_{v \in V} \| f_G(v) - f_H(\varphi'(v)) \| = \max_{w\in \rho (V)} \| f_{G'}(w) - f_H(\psi'(w)) \| \\
    &\geq \inf_{\psi\in \mathfrak{G}'}  \max_{w \in \rho(V)} \| f_{G'}(w) - f_{H'}(\psi(w)) \| = \dis(G',H').
\end{align*}

 In a similar manner we can reason 
    \begin{align*}
     \dis(H,G)\geq  \dis(H',G').
\end{align*}
From which it follows 
\begin{align*}
    \dhd{G}{H}\geq \dhd{G'}{H'}.
\end{align*}
With an identical argument as above we have for any $\psi\in \mathfrak{G}' \text{ there exists a map }
\varphi= \gamma^{-1}\circ \psi\circ \rho\in \mathfrak{G} $ such that $  \max_{v \in V} \| f_G(v) - f_H(\varphi  (v))\|=\max_{w \in \rho(V)} \| f_{G'}(w) - f_{H'}(\psi(w)) \| $. Thus, we can deduce 
\begin{align*}
      \dhd{G'}{H'}\geq \dhd{G}{H}.
\end{align*}
To which we may conclude, $  \dhd{G}{H} = \dhd{G'}{H'}.$
\end{proof}

\psipower*
\begin{proof}
 Let $\mathfrak{G}=\Hom{G}{H}\text{ and }\mathfrak{G}'=\Hom{H}{G} $. We proceed case by case:
\begin{enumerate}
    \item  
Assume $\dhd{G}{H}=0\implies \max\{\dis(G,H),\dis(H,G)\}=0$.\\
    Since $\max\{\dis(G,H),\dis(H,G)\}=0 $ we have $ \mathfrak{G},\mathfrak{G}'\neq \emptyset \implies \textrm{core}(G) \simeq \textrm{core}(H).$\\\\
Assume $\textrm{core}(G) \simeq \textrm{core}(H) \implies  \mathfrak{G},\mathfrak{G}'\neq \emptyset $. For any $\psi\in \mathfrak{G}$ \begin{align*}
        \max_{v \in V} \|f_G(v) - f_H(\psi(v))\|= \max_{v \in V} \|\Psi \circ \iota_G(v) - \Psi \circ \iota_H(\psi(v))\|= 0   \quad (\Psi \text{ is constant)}
    \end{align*}
    Since $\mathfrak{G} \neq \emptyset$ we have  $\dis(G,H)=0$. Similarly we can conclude $\dis(H,G)=0$. Therefore $\dhd{G}{H}=0.$

\item     Suppose $\dhd{G}{H}=0$ and assume to the contrary that $G\not\simeq H$. \\
    Case 1: $\mathrm{core}(G)\not\simeq \mathrm{core}(H)$. Without loss of generality assume $\mathfrak{G}= \emptyset$. This would then give us $\dis(G,H)=\infty\implies\dhd{G}{H}=\infty\neq 0=\dhd{G}{H}$. This is a contradiction.  \\
Case 2: $\mathrm{core}(G)\simeq \mathrm{core}(H) \implies\mathfrak{G},\mathfrak{G}'\neq \emptyset$. For any $\psi\in \mathfrak{G}$ and any vertex $v\in V(G)$ we have $f_G(v) =\Psi\circ \iota_G(v)$. Since $G\not\simeq H$ it follows that there does not exist any rooted isomorphism between $(G,v)\text{ and } (H,\psi(v))$. Thus $\iota_G(v)\neq \iota_H(\psi(v))$ and since $\Psi$ is injective we have $f_G(v)=\Psi\circ \iota_G\neq \Psi\circ\iota_H(\psi(v))= f_H(\psi(v))$. Putting this together 
\begin{align*}
   \forall\psi\in \mathfrak{G},\quad \max_{v\in V(G)}\|f_G(v)-f_H(\psi(v))\|> 0\implies \dis(G,H)> 0.
\end{align*}
Since $\mathfrak{G}'\neq \emptyset$ we can similarly reason $ \dis(H,G)>0$. This gives us the contradiction $\dhd{G}{H}>0$.\\\\
On the other hand if we assume $G\simeq H$ by an isomorphism $\psi\in \mathfrak{G}$. For any vertex $v\in V(G),\;f_G(v)=\Psi\circ \iota_G(v)=\Psi\circ \iota_H(\psi(v))=f_H(\psi(v))$. Putting this together 
\begin{align*}
    0\leq \dis(G,H)\leq  \|f_G(v) - f_H(\psi(v))\|=0\implies \dis(G,H)=0.
\end{align*} There exists an isomorphism $\varphi\in \mathfrak{G}'$ hence, in a similar manner we can reason $\dis(H,G)=0$. Therefore $\dhd{G}{H}=\max\{\dis(G,H),\dis(H,G)\}=0$. 

\item From  \Cref{prop:perm_inv} the metric $\hd$  is well defined on the quotient. $\dhd{(G_1,v_1)}{(H_1,w_1)}=\dhd{(G_2,v_2)}{(H_2,w_2)}= \dhd{[(G,v)]}{[(H,w)]}$, for any $(G_1,v_1),(G_2,v_2)\in[(G,v)] $ and $ (H_1,w_1),(H_2,w_2)\in[(H,w)]$.  $((\mathcal{G_{\bullet}}/\simeq_{\bullet}),\hd)$ inherits non-negativity, symmetry, and the triangle inequality directly from the metric $(\mathcal{G_{\bullet}},\hd)$. Additionally 2. describes non-degeneracy confirming $((\mathcal{G_{\bullet}}/\simeq_{\bullet}),\hd)$ is a metric space.

\end{enumerate}
\end{proof}

\subgraphdist*
\begin{proof}
    We start with $\dhd{G'}{H'}$ where by triangle inequality
    \begin{equation*}
        \dhd{G'}{H'} \leq \dhd{G'}{G} +  \dhd{G}{H'}
    \end{equation*}
    then 
    $$
    \dhd{G}{H'} \leq \dhd{G}{H} + \dhd{H}{H'}
    $$
    combine together and we get that
    \begin{equation*}
        \dhd{G'}{H'} \leq \dhd{G}{G'} + \dhd{G}{H} + \dhd{H}{H'}
    \end{equation*}
    we move the distances to the subgraphs to the left hand  side
    \begin{equation*}
        \dhd{G'}{H'} - \left( \dhd{G}{G'} +   \dhd{H}{H'} \right ) \leq  \dhd{G}{H},
    \end{equation*}
    then the claim follows if we flip arround the inequality.
\end{proof}

\section{Proofs of Results on Distinction Capabilities}\label{app:proofs_disct}

\begin{restatable}[Equal colors under WL coloring equivalence]{lemma}{wlhisto}
    \label{WL_histogram}
    Let $G$ and $G'$ be two graphs of order $n$ whose attribute functions $c$ and $c'$ correspond to their respective WL colorings. Assuming identical color sets $\mathcal{C} = \mathcal{C}'$, let $N_k(G)$ denote the number of nodes in $G$ with color $k$. Then, 
$N_k(G) = N_k(G')$ for each color $k \in \mathcal{C}$.
\end{restatable}
\begin{proof}
    The stable coloring of nodes obtained through the WL algorithm results in a partition of vertices defined by their colors: $\{V_1, ... V_m\}$ which is a \textit{equitable partition}, which means that, $\forall v_i \in V_i, d_{ij} := |\{u \in \mathcal{N}_g(V) : c(u) = j\}|$ is fixed. This means that the amount of edges between nodes of color $i$ and color $j$ is fixed and strictly determined by the colors themselves. This implies that, given a node of color $i$, all the values $d_{i,j}$ are uniquely defined for each $j$, which also implies that its neighborhood structure is fixed, in terms of the amount of neighboring nodes of each color. \\
    This means though that both $G$ and $G'$ satisfy the following system of equations in terms of the number of their nodes belonging to the different classes, $\mathbf{N} = [N_1, ..., N_m]$:
    \begin{equation}\label{system}
    \begin{cases}
    N_i d_{i,j} = N_j d_{j,i} \quad \forall i, j\\
    \sum_i N_i = n
    \end{cases}
    \end{equation}
    \\
    We now demonstrate that the solution $\mathbf{N}$ to this system is unique by (1) showing that it is unique up to a scalar factor, by looking at the first equation and analyzing the associated transition matrix, and (2) exploiting the second equation to obtain the actual uniqueness.
    \\
    Let $D_i = \sum_{j} d_{i,j}$ be the total degree of color class $i$. We define a matrix $T$ of size $m \times m$ describing the random walk on the color classes:
    $ T_{i,j} = \frac{d_{i,j}}{D_i}, $ which is a row-stochastic transition matrix, i.e., rows express the probabilities of jumping from one color class to all other, through edge connections.
    \\
    We introduce a variable $\boldsymbol{\pi}$ to represent the total volume of each class (the total number of edges from such class to others), which will represent the stationary distribution of our discrete Markov chain. In particular, let $\pi_i = N_i D_i$. Substituting $d_{i,j} = T_{i,j} D_i$ into the detailed balance equation, i.e., the first row of \Cref{system}, we obtain:
    $$ N_i (T_{i,j} D_i) = N_j d_{j,i} \implies \pi_i T_{i,j} = N_j d_{j,i}. $$
    Summing this equation over all possible starting classes $i$ yields the global balance for class $j$:
    \begin{equation}
        \sum_{i=1}^m \pi_i T_{i,j} = N_j \sum_{i=1}^m d_{j,i} = N_j D_j = \pi_j.
    \end{equation}
    In vector notation, this is equivalent to:
    \begin{equation} \label{eq:eigenvector}
        \boldsymbol{\pi} T = \boldsymbol{\pi}.
    \end{equation}
    \Cref{eq:eigenvector} identifies the vector $\boldsymbol{\pi}$ as the left eigenvector of the transition matrix $T$ corresponding to the eigenvalue $\lambda = 1$.
    \\
    Since $G$ and $G'$ are assumed to be connected, the induced quotient graph is strongly connected, which in matrix terms implies that $T$ is irreducible. The reason why this is relevant is that, according to the Perron-Frobenius Theorem, an irreducible stochastic matrix $T$ has a unique 1-eigenvector, hence $\boldsymbol{\pi}$ is unique up to scalar normalization. 
    Since $\pi_i = N_i D_i$ and the degrees $D_i$ are fixed structural constants identical for both graphs, the uniqueness of $\boldsymbol{\pi}$ implies the uniqueness of the frequency vector $\mathbf{N}$ up to a scalar factor $\alpha$:
    $$ \mathbf{N} = \alpha \cdot \mathbf{v}_{base}, $$
    where $\mathbf{v}_{base}$ is a fixed reference vector determined solely by the matrix $T$. This directly implies that also the solution to the first equation in \Cref{system} is unique.
    \\
    Finally, the normalization constraint $\sum N_i = n$ uniquely determines the scalar $\alpha$:
    $$ \alpha = \frac{n}{\sum_{i} (\mathbf{v}_{base})_i} $$
    Since the transition matrix $T$ (derived from $d_{i,j}$) and the total count $n$ are identical for both graphs, the frequency vector $\mathbf{N}$ is uniquely determined. Therefore:
    $$ N_k(G) = N_k(G') \quad \forall k \in \mathcal{C}. $$
\end{proof}

\hdwl*
\begin{proof}
     Since we are analyzing the WL isomorphism test between two graphs, we can assume that their cardinality is the same, that is, $|V| = |V'| = n$. Let $G$ and $H$ be two non-isomorphic graphs, and suppose WL can distinguish them. We can then consider two cases:
    \begin{enumerate}
        \item Suppose $\mathfrak{G} = \Hom{G}{H} = \emptyset$ or $\mathfrak{H} = \Hom{H}{G} = \emptyset$. Then, the graphs are not isomorphic, and we have $$\mathfrak{G} = \emptyset \Rightarrow \inf_{\varphi \in \mathfrak{G}} = \infty,$$ $$\mathfrak{H} = \emptyset \Rightarrow \inf_{\psi \in \mathfrak{H}} = \infty,$$ which imply $\dhd{G}{H} = \infty > 0$.
        \item Suppose $\mathfrak{G} \neq \emptyset$ and $\mathfrak{H} \neq \emptyset$, and define the final graph coloring (obtained through the color refinement algorithm) as integers features, e.g. $c_i = i$, such that 
        $$\| c(v) - c'(\varphi(v)) \| \iff c(v) \neq c'(\varphi(v)).$$
        Suppose now that $\dhd{G}{H} = 0$. Then,
        $$\exists \varphi \in \mathfrak{G} : \forall v \in V(G) \quad \| c(v) - c'(\varphi(v)) \| = 0,$$
        $$\exists \psi \in \mathfrak{H} : \forall w \in V(H)  \quad \| c'(w) - c(\psi(w)) \| = 0$$
        Which implies that
        $$\exists \varphi \in \mathfrak{G} : \forall v \in V(G) \quad c(v) = c'(\varphi(v)),$$
        $$\exists \psi \in \mathfrak{H} : \forall w \in V(H)  \quad c'(w) = c(\psi(w)).$$
        Define now $\mathcal{C}(V) :=\text{ set of colors on } V(G)$. Then two previous two equations imply, respectively, $$\mathcal{C}(V(G)) \subseteq \mathcal{C}(V(H)),$$ $$\mathcal{C}(V(H)) \subseteq \mathcal{C}(V(G)),$$ which combined imply $\mathcal{C}(V) = \mathcal{C}(V(H))$. Now, since the set of colors appearing on the two graphs are the same, the only way for the WL test to result in distinguishing them is that their histogram should be different, that is, that there exist at least one color  the amount of the appearance of each color in the two graphs differ. To elaborate on this point, let's define, for each color $k \in \mathbb{N}$, $N_k(G)$ as the amount of nodes in $G$ with color $k$. So, the color histogram for graph $G$ is defined by $[N_1(G), N_2(G), ... N_K(G)]$, with $K = |\mathcal{C}(V(G))| = |\mathcal{C}(V(H))|$, and the one for $H$ is defined by $[N_1(H), N_2(H), ... N_K(H)]$. At this point, we can exploit \Cref{WL_histogram}, obtaining the result that the histograms for $G$ and $H$ are the same. This would lead to WL not distinguishing the two graphs, which leads to a contradiction with the initial assumptions. This implies that the assumption $\dhd{G}{H} = 0$ cannot hold, hence $\dhd{G}{H} > 0$.
    \end{enumerate}
\end{proof}

\discpowercolor*
\begin{proof}
    We can address the two inequalities in order: (1) first of all, we will show $\dwl{G}{H} \leq \ddh(G, H)$, and then (2) $\ddh(G, H) \leq \dhd{G}{H}$.
    \begin{enumerate}
        \item If the histograms of the graphs are equal $\mathrm{h}(G) = \mathrm{h}(H)$, then both $\wl$ and $\ddh$ are 0, and because $0 \leq 0$, the inequality holds. In case the histograms differ, we will have $\dwl{G}{H} = 1$. Additionally, because the histograms of $G$ and $H$ differ, from \Cref{WL_histogram} there will be at least one color $c$ that appears in only one between $h(G)$ and $h(H)$. The existence of such color will imply by definition that $\ddh(G, H) \geq 1$, leading to $\dwl{G}{H} \leq \ddh(G, H)$.
        \item Consider now the second inequality: $\ddh(G, G') \leq \dhd{G}{H}$. Suppose that $\ddh(G, H) =: \alpha$ and, without loss of generality, suppose that this value of the distance is given by a specific pair of nodes $v \in G \quad \mathrm{and} \quad w \in H$, i.e., $\alpha = \|c(v) - c'(w)\|$. But also the following set of equalities holds, given a specific $\varphi$:
        \begin{align*}
        \ddh(G, H)  = \\
        & = \alpha \\
        &= \|c(v) - c'(w)\| \\
        &= \min_{\hat{v} \in H} \|c(v) - c'(\hat{v})\|\\
        & \leq \inf_{\varphi \in \mathfrak{G}} \|c(v) - c'(\varphi(v))\|\\
        & \leq \inf_{\varphi \in \mathfrak{G}} \{\max_{v \in G} \|c(v) - c'(\varphi(v))\|\}\\
        & = \mathrm{dis}(G, H) \\
        & \leq \dhd{G}{H}
        \end{align*}
    which proves the statement.
    \end{enumerate}

\end{proof}

\metrictopol*
\begin{proof}
   
    We proceed to show the containment  $\mathcal{T}_{\mathrm{WL}}\subseteq \mathcal{T}_{\mathrm{HD}}$ by the basis characterization \begin{align*}
        \mathcal{T}_{\mathrm{WL}}\subseteq \mathcal{T}_{\mathrm{HD}} \iff\forall G\in \mathcal{ G}\text{ and }G\in B\in \mathcal{B}_{WL},\;\exists B'\in \mathcal{B}_{HD} \text{ such that }  G\in B'\subseteq B. 
    \end{align*}
For any $G\in \mathcal{G}$ the  elements of $\mathcal{B}_{\mathrm{WL}}$ are defined as the open balls w.r.t. metric $\wl$ , thus denote the open ball of radius $r>0$ centered at $G$ by $B= B_{\mathrm{WL}}(G,r)$.
   \begin{enumerate}
       \item If $r>1.$ We have $ B= B_{\mathrm{WL}}(G,r) =\mathcal{G}$. Set \begin{align*}
           r'=\max_{H\in \mathcal{G}}\{\dhd{G}{H}\}+\epsilon\quad,\text{ for } \epsilon>0
       \end{align*} then $ B'=B_{\mathrm{HD}}(G,r')=\mathcal{G}$. Thus there exists $B'\in \mathcal{B}_{\mathrm{HD}}$ such that $G\in B'\subseteq B$.
       \item If $r<1.$ We have $B=B_{\mathrm{WL}}(G,r) =\{H\in \mathcal{G}|\; \wl(G,H)=0\}$. Set  \begin{align*}
           r'=\min_{H\in \mathcal{G}}\{\dhd{G}{H}\}
       \end{align*} then $B'=B_{\mathrm{HD}}(G,r')=\{H\in \mathcal{G|\;}\dhd{G}{H}=0\}$. We have $G\in B' \text{ and } B'\in \mathcal{B}_{\mathrm{HD}}$. For any graph $H\in B'$ from \cref{theo:discpower2} we have $0\leq \wl{G}{H}\leq \dhd{G}{H}=0$ thus $\wl(G,H)=0$ and  $H\in B$ therefore $B'\subseteq B$ . There exists $B\in \mathcal{B}_{\mathrm{HD}}$ such that $G\in B'\subseteq B$. 
       
   \end{enumerate}
From 1. and 2. it follows that $\mathcal{T}_{\mathrm{WL}}\subseteq \mathcal{T}_{\mathrm{HD}}$. 
To show the strict containment consider the Shrikhande graph $S$ and the $4\times 4$ Rook's  graph $R$ \citep{arvind2020weisfeiler}. In \citet{arvind2020weisfeiler} authors show that for $S$ and $R$ color histograms $h(S)=h(G)$ and they are not homomorphic since $S$ contains a $4$-clique and $S$ does not. Thus, $\wl(S,R)=0$. For any open set $ U\in \mathcal{T}_{\mathrm{WL}}$  where $R\in U$ we can decompose $U$ into its basis elements \begin{align*}
    U= \bigcup_{i\in I} B_i
\end{align*}
here $B_i$ are all basis elements and $I$ is some indexing set. Thus for some graph $G\in \mathcal{G}$ with radius $r'>0$ and index $\lambda\in I$ we have $R\in B_{\lambda}=B_{\mathrm{WL}}(G,r')$. Since $\wl(S,R)=0$ it follows $S\in B_{\lambda}\subseteq U$. Thus any open set in $\mathcal{T}_{\mathrm{WL}}$ which contains $R$ contains $S$. On the other hand since
\begin{align*}
    \Hom{R}{S}=\emptyset\implies \dis(R,S)=\infty\implies \dhd{R}{S}=\infty.
\end{align*}
For any fixed $0<r<\infty$ consider the open set $B= B_{\mathrm{HD}}(R,r)\in \mathcal{T}_{\mathrm{HD}}$. Note $R\in B \text{ and }S\not\in B$ therefore $B\not\in \mathcal{T}_{\mathrm{WL}}$. We can then conclude $\mathcal{T}_{\mathrm{WL}} \subset \mathcal{T}_{\mathrm{HD}}$. (QED)

\end{proof}

\infdis*
\begin{proof}
    We proof the statements one at a time:
    \begin{enumerate}
        \item 
    This is a rather simple case and has a name. The function $\varphi'\colon G' \to G$ we are looking for is the \emph{inclusion homomorphism}, where for every $v \in V'$ we have $\varphi'(v) = v$. $\varphi'$ is a homomorphism since by definition of an induced subgraph if $\{u,v\} \in E'$ then it's also in $E'$. Note now that for $\varphi'$, we have 
    \begin{align*}
        \max_{v' \in V'} \|f_{G'}(v') - f_G(\varphi'(v'))\| 
        &= \max_{v' \in V'} \|f_{G'}(v') - f_{G'}(\varphi'(v))\|  \\
        &= \max_{v' \in V'} \|f_{G'}(v') - f_{G}(v)\|  \\
        &= \max_{v' \in V'} 0 \\
        &= 0
    \end{align*}
    \item  A retraction $r\colon G \to G'$ is a homomorphism that always satisfies $r(v) = v$. Note that to obtain a $0$ we need a retraction $r\colon G \to G'$. Since for $K_n$ and $C_{\ell}$ of odd $\ell$ only admit a retraction to themselves \citep{hell2004graphs} and $G \neq G'$, the claim follows. 
    \end{enumerate}

\end{proof}

\embeddinginvariance*
\begin{proof}
    Let $\rho \colon V \to V$ be an isomorphism on $G$ and let $H= (\rho(V), E', f_H)$ be the isomorphic graph resulting from applying the isomorphism, where $E'$ is induced by permuting vertices $V$. We aim to show $\Gamma(G)_{\mathcal{F}}=\Gamma(H)_{\mathcal{F}}$. This equivalent to showing
    \begin{align*}
        [\dhd{F}{G}]_{F\in \mathcal{F}}= [\dhd{F}{H}]_{F\in \mathcal{F}}
    \end{align*}
Since $H$ is isomorphic to $G$ from  \Cref{prop:perm_inv} it follows that for any fixed $F\in \mathcal{F}$ we have $\dhd{F}{G}=\dhd{F}{H}$. With this fact it follows that $\Gamma(G)_{\mathcal{F}}=\Gamma(H)_{\mathcal{F}}$.

\end{proof}

\embeddingcomplete*
\begin{proof}
    Let $G,H \in \mathcal{G}_n$ be a graph with attribute functions $f_G\text{ and }f_H$ respectively.  We want to show that $G \simeq H$ if and only if $\hde{G} = \hde{H}$.\Cref{lemma:hom_enc_invariant} establishes the sufficiency of this condition. We need only to verify the necessity. The distinguishing power of $f_G \text{ and }f_H$ are determined by $\Psi$ hence, assume $\Psi$ distinguishes $G,H$, i.e., $\ddh(V(G), V(H)) > 0$, and $\hde{G} = \hde{H}\implies $ for any fixed graph $F\in \mathcal{F},\; \dhd{F}{G}= \dhd{F}{H}$ . Since $\mathcal{F}= \mathcal{G}_n$ it follows that $G,H\in  \mathcal{F}$ thus for $F=H\text{ we have },\;0=\dhd{H}{H}=\dhd{H}{G}$. From \Cref{coro:psipower} we have $G\simeq H$.

\end{proof}

\section{Further Experimental Details}\label{app:extended_experiments}

We provide brief details on our experiments, described in \Cref{sec:results} in the main text.

\paragraph{Choices of $\mathcal{F}$.}
The family of cycle graphs $\mathcal{F}_{n}^{C}$ contains all cycle graphs up to $C_{n}$. Since the smallest cycle graph possible is $C_{3}$, then $|\mathcal{F}_{n}^{C}| = n - 2$. \paragraph{Treewidth statistics.}
\Cref{tbl:treewidths_ds} shows the distribution of treewidths along with the number of graphs and the maximum number of nodes, all elements required to measure the computational cost of running our method.

\begin{wraptable}{r}{0.5\textwidth}
\caption{Distribution of treewidths (show in $\%$) per dataset}
\label{tbl:treewidths_ds}
\sisetup{detect-all
=true,detect-weight=true,table-align-uncertainty=true,table-comparator=true,table-format=<2.3(3.1),tight-spacing=true}
\centering
    \begin{small}\begin{tabular}{@{}l@{\hspace{8pt}} S[table-format=2.3] S[table-format=2.3] S[table-format=2.3]}
        \toprule
        Property & \texttt{ZINC} & \texttt{NCI1} & \texttt{ogbg-molhiv} \\
        \midrule
        $\mathrm{tw}(G) = 1$ & 0.550  & 2.019  & 3.847 \\
        $\mathrm{tw}(G) = 2$ & 99.033 & 92.360 & 92.023 \\
        $\mathrm{tw}(G) = 3$ & 0.417  & 5.572  & 4.063 \\
        $\mathrm{tw}(G) = 4$ & 0.000  & 0.024  & 0.063 \\
        $\mathrm{tw}(G) = 5$ & 0.000  & 0.000  & 0.002 \\
        \midrule
        Max $\mathrm{tw}(G)$ & 3  & 4   & 5 \\
        Max $|V(G)|$         & 37 & 111 & 222 \\
        \bottomrule
        \end{tabular}
    \end{small}\end{wraptable} 

\paragraph{Algorithmic computations}

We provide the pseudocode for the implementation of the calculation of the homomorphism distortion. Primarily, this algorithm relies on a \textsc{NiceTreewidthDecomposition}, we then perform a dynamic programming technique from \citet{hutchison_fine_2013}, originally proposed for the ``Minimum Weighted Homomorphism Problem". the decomposition to build a DP table that will allow use to traverse the search space using a min-max approach. 
The solution relies on the construction of a DP table and a traversing it to solve a min-max problem by accounting for partial homomorphism from each of the vertices in a bag.
Our implementation of the tree decomposition relies on finding an initial good approximation of treewidth with the \textsc{networkx.treewidth\_min\_degree} function from library \textsc{networkx}\citep{hagberg2020networkx}, then refining said approximation to the optimal treewidth by an interpretation of the algorithmic procedure described in \citep{kloks1994treewidth} and finally performing \cref{alg:mindistortion} process as a dynamic programming interpretation from \citet{hutchison_fine_2013}.

\begin{algorithm}[t]
\caption{\textsc{MinDistortionHomDP}$(G, H, f_G, f_H)$ --- min--max
distortion over $\mathrm{Hom}(G,H)$ via a nice tree decomposition of $G$}
\label{alg:mindistortion}
\begin{algorithmic}[1]
\Require Graphs $G,H$, features $f_G,f_H$ 

\State $\delta(v,w) \gets \lVert f_G(v)-f_H(w)\rVert$ \Comment{precompute all-pairs distance matrix}
\State $N_H(w) \gets \{w' : (w,w')\in E(H)\} \cup \{w : (w,w)\in E(H)\}$ for each $w\in V(H)$
\State $(T, \{X_t\}, r) \gets$ \Call{NiceTreeDecomposition}{$G$} \Comment{$T$ is the tree, $X_t$ is the bag at node $t$ and $r$ is the root}
\State $c_t \gets \{\}$ for all $t \in T$ \Comment{DP tables of $\phi: X_t \to V(H)$, where each key is $\{(u  ,w)  \mid u \in X_t \text{ and } \in V(H)) \text{ such that there exist a homomorphism } \phi(u) = w\}$ }

\For{$t \in$ \Call{PostOrder}{$T, r$}} \Comment{\textsc{PostOrder} returns a stack that guarantees that when a node is seen all of its children have been already seen.}
  \If{$\mathrm{kind}(t) = \textsc{leaf}$}
    \State $c_t(\varnothing) \gets 0$

  \ElsIf{$\mathrm{kind}(t) = \textsc{introduce}(v)$, child $t'$}
    \For{$\phi' \in \operatorname{dom}(c_{t'})$}  \Comment{Each $\phi \in \mathrm{dom}(c_{t'})$ is a pair $(u \in X_t, w\in V(H))$ where $\phi(u) = w$ a homomorphism.}
      \For{$w \in V(H)$}
        \State $\mathrm{ok} \gets \textbf{true}$
        \For{$u \in X_t \setminus \{v\}$}
          \If{$(u,v)\in E(G)$ \textbf{and} $\phi'(u) \notin N_H(w)$}
            \State $\mathrm{ok} \gets \textbf{false}$
          \EndIf
        \EndFor
        \If{$\mathrm{ok}$}
          \State $\phi \gets \phi' \cup \{v \mapsto w\}$ \Comment{This is the pair $(u,w)$}
          \State $c_t(\phi) \gets \min\bigl(c_t(\phi),\; \max(c_{t'}(\phi'),\, \delta(v,w))\bigr)$
        \EndIf
      \EndFor
    \EndFor

  \ElsIf{$\mathrm{kind}(t) = \textsc{forget}(v)$, child $t'$}
    \For{$\phi \in \operatorname{dom}(c_{t'})$}
      \State $\phi_{-v} \gets \phi \setminus \{v \mapsto \phi(v)\}$
      \State $c_t(\phi_{-v}) \gets \min\bigl(c_t(\phi_{-v}),\; c_{t'}(\phi)\bigr)$
    \EndFor

  \Else \Comment{\textsc{join}, children $t_1, t_2$}
    \For{$\phi \in \operatorname{dom}(c_{t_1})$}
      \If{$\phi \in \operatorname{dom}(c_{t_2})$}
        \State $c_t(\phi) \gets \max\bigl(c_{t_1}(\phi),\, c_{t_2}(\phi)\bigr)$
        \Comment{max since $V_{t_1}\cap V_{t_2}=\varnothing$}
      \EndIf
    \EndFor
  \EndIf
\EndFor

\If{$c_r = \varnothing$}
  \State \Return $+\infty$ \Comment{no homomorphism exists}
\EndIf
\State \Return $\min_{\phi \in \operatorname{dom}(c_r)} c_r(\phi)$
\end{algorithmic}
\end{algorithm}

\paragraph{Computational Complexity}

Given two graphs $G = (V,E), H=(W,E')$ and $|V|=n, |W|=m$, let $\mathrm{tw}(G) = k$. We present the computational complexity of finding out $k$, the tree decomposition for $k$ and performing the DP.
First, assume we do not know $k$, the algorithm in \textsc{networkx} gives an approximate treewidth $k'$ such that $k \leq k'$ in $\gO(V^3)$ time. Then, \citet{kloks1994treewidth} gives a $\gO(k \cdot n)$ time algorithm for constructing the final \emph{nice} tree decomposition.
Finally, the complexity of the DP is $\gO(m^{k+1}\cdot k^2 \log k \cdot n)$ by \citep{hutchison_fine_2013}, where $\gO(k \log k)$ is the amount of nodes in the tree, $\gO(k \log k)$ is the sorting of processing the DP table at certain nodes an $m \coloneq q$ in the reference.

\paragraph{Hardware setup. }

For $\texttt{BREC}$ experiments and precomputation of the homomorphism encoding we use a compute node with $\times 2$ \textsc{AMD EPYC 9554 64-core}.  For model training, we use $\times4$ \textsc{AMD EPYC 9654 96-Core Processor} and $\times 8$ \textsc{NVIDIA RTX PRO 6000 Blackwell Server Edition}  with \texttt{CUDA} version 13.0, and Pytorch version 2.11.0.

\paragraph{Hyperparameters.}
\Cref{tbl:hyperparams_encoding} shows the hyperparameters of the homomorphism distortion encoding. \Cref{tab:fixed_hyperparams} shows the fixed hyperparamters shared accross all the training setup, \Cref{tab:gcn_gat_hyperparams} shows the hyperparamters for the GCN and GAT models when run on \texttt{ZINC} and \Cref{tab:gin-hyperparams} shows the hyperparamters for the GIN model for \texttt{ZINC}, \texttt{ogbg-molhiv} and \texttt{NCI1}.

\begin{table}[tbp]
    \centering  
    \caption{Encoding Hyperparamters.}
    \label{tbl:hyperparams_encoding}
    \begin{tabular}{l c }
        \toprule
        Hyperparameter & Values  \\
        \midrule
        Walk length &  $\{10, 75, 150, 200\}$ \\
        $\mathrm{tw}$ & $\{1, 2\}$ \\
        $\mathcal{F}^C$ &  $\{C_8, C_9\}$ \\

        \bottomrule
    \end{tabular}
\end{table}

\begin{table}[t]
\centering
\caption{Fixed hyperparameters.}
\label{tab:fixed_hyperparams}
\begin{tabular}{lc}
    \toprule
    Parameter & Value \\
    \midrule
    \quad Residual & True \\
    \quad Normalization    & Batch  \\
    \quad Initial Lr& 0.001 \\
    \quad Batch Size & 128   \\
    \quad Min. LR   & $10^{-5}$ \\
    \bottomrule
\end{tabular}
\end{table}

\begin{table}[t]
\centering
\caption{Hyperparameters for GCN and GAT on \texttt{ZINC}.}
\label{tab:gcn_gat_hyperparams}
\begin{tabular}{lcc}
    \toprule
    Parameter & GCN & GAT \\
    \midrule
    \multicolumn{3}{l}{\textit{Architecture}} \\
    \quad Layer    & 16   & 16 \\
    \quad Hidden Dimension    & 172  & 22 \\
    \quad Hidden Output & --   & 176 \\
    \quad Heads     & --   & 8 \\
    \quad Input Dim     & 28   & 28 \\
    \quad Readout        & Mean & Mean \\
    \quad Residual       & True & True \\
    \quad Normalization    & Batch & Batch \\
    \quad Dropout        & $\{0.0, 0.5\}$  & -- \\
    \midrule
    \multicolumn{3}{l}{\textit{Training}} \\
    \quad Max Epoch      & 1000              & 1000 \\
    \bottomrule
\end{tabular}
\end{table}

\begin{table}[t]
\centering
\caption{Hyperparameters on GIN in \texttt{ZINC}, \texttt{ogbg-molhiv}, and \texttt{NCI1}.}
\label{tab:gin-hyperparams}
\begin{tabular}{lccc}
    \toprule
    Parameter & \texttt{ZINC} & \texttt{ogbg-molhiv} & \texttt{NCI1} \\
    \midrule
    \multicolumn{4}{l}{\textit{Architecture}} \\
    \quad Layers        & 16    & 5                        & 5 \\
    \quad Hidden Dimension         & 124   & 300                      & 64 \\
    \quad Input Dimension          & 28    & 300  & 37 \\
    \quad Readout           & Sum   & Sum                      & Sum \\
    \quad Dropout             & $\{0.0, 0.5\}$   & $\{0.0, 0.5\}$    & $\{0.0, 0.5\}$\\
    \midrule
    \multicolumn{4}{l}{\textit{Training}} \\
    \quad Max Epoch      & 1000                & 150 & 350 \\
    \bottomrule
    \end{tabular}    
\end{table}

\end{document}

%% file: figures/intro_fig_tikz.tex
\begin{tikzpicture}[xscale = 1.25]
\definecolor{setA}  {RGB}{127,191,123}
\definecolor{setB}  {RGB}{175,141,195}

\definecolor{colHD} {RGB}{103,169,207}
\definecolor{colH}  {RGB}{239,138,98}

% ---------------------------------------------------------------------------
% Styles
% ---------------------------------------------------------------------------
\tikzset{
  % attributes: circle = attribute of G, square = attribute of H
  gpt/.style 2 args = {circle, fill = #1, draw = #2, inner sep = 0.1pt, minimum size = 6.5pt, font=\large},
  hpt/.style 2 args = {circle, fill = #1, draw = #2, inner sep = 0.1pt, minimum size = 6.5pt, font=\large},
  % graph vertices
  gv/.style    = {circle, fill = #1, inner sep = 0pt, minimum size = 3.6pt},
  hv/.style    = {circle, fill = #1, inner sep = 0pt, minimum size = 3.6pt},
  gedge/.style = {line width = 0.6pt},
  stem/.style  = {densely dashed, black!35, line width = 0.4pt},
  % matchings (coloured by the source vertex)
  match/.style = {-, draw = #1, line width = 0.3pt,
                  shorten <= 1.5pt, shorten >= 1.5pt},
  best/.style  = {match = #1, line width = 1.2pt},
  nnbad/.style = {match = black!40, densely dotted, line width = 0.6pt},
  lab/.style   = { inner sep = 2.2pt},
  % --- decorations ---------------------------------------------------------
  % coloured box around a whole row + rotated tag sitting on its left edge
  rowbox/.style  = {rounded corners = 4pt, fill = #1!7, draw = #1!70,
                    line width = 0.5pt},
  rowtag/.style  = {rotate = 90, fill=white, draw = #1, text = black,
                    rounded corners = 2pt, inner sep = 2pt,
                    minimum width = 1.6cm},
  % shaded bands marking the layer of graph G / graph H (bottom row)
  band/.style    = {loosely dashed, fill = #1, fill opacity=0.1, rounded corners = 3pt, draw=black, line width = 0.1pt, inner sep=3pt},
  bandlab/.style = {text = #1, inner sep = 1pt},
  % separator between the two rows
  divider/.style = {densely dashed, black!45, line width = 0.5pt},
  % "= max ( . , . )" wrapped around each row
  eqlab/.style    = { font = \normalsize, anchor = east, inner sep = 1pt},
  maxparen/.style = {font = \normalsize, inner sep = 0pt, black!85},
bigcomma/.style = {font = \Huge, scale = 1.37, anchor = base,
                   inner sep = 0pt, black!85},
    % direction of the homomorphism (G -> H or H -> G)
  homarrow/.style = {-{Stealth[length = 5pt, width = 4pt]}, line width = 0.9pt,
                     line cap = round, dash pattern = on 0pt off 2.4pt,
                     black!70, shorten <= 2pt, shorten >= 1pt},
  homlab/.style   = {font = \normalsize, inner sep = 2pt},
}

% ---------------------------------------------------------------------------
% Horizontal layout (unscaled cm).  Each X box spans x = shift+1.15 ... shift+6.9.
% The G / H labels + homomorphism arrow sit left of the left X box (phi) and
% right of the right X box (psi); only the big comma sits between the panels.
% ---------------------------------------------------------------------------
\newcommand{\panelL}{1.85}  % xshift of the left  panels  (X box: 3.00 ... 8.75)
\newcommand{\panelR}{8.45}  % xshift of the right panels  (X box: 9.60 ... 15.35)
\newcommand{\rowL}{0.5}     % left  edge of the row boxes (tag is centred here)
\newcommand{\rowR}{17.2}    % right edge of the row boxes
\newcommand{\parenL}{1.9}   % left edge of the opening parenthesis
\newcommand{\parenR}{16.65} % left edge of the closing parenthesis
\newcommand{\bandL}{2.3}    % left  edge of the G / H bands
\newcommand{\bandR}{16.6}   % right edge of the G / H bands

% Tall parentheses typeset by TeX (same glyphs as \left( ... \right)).
%   #1 = x of the parenthesis' left edge, #2 = y of its centre, #3 = height
\newcommand{\openparen}[3]{%
  \node[maxparen, anchor = west] at (#1,#2)
    {$\left(\vcenter{\hbox{\rule{0pt}{#3}}}\right.$};}
\newcommand{\closeparen}[3]{%
  \node[maxparen, anchor = west] at (#1,#2)
    {$\left.\vcenter{\hbox{\rule{0pt}{#3}}}\right)$};}

% ---------------------------------------------------------------------------
% The attribute space X = R^2 (flat).  Attributes f_G(v_i) = x_i, f_H(w_j) = y_j.
% Positions chosen so that (Euclidean distance in the drawing):
%   G -> H: nearest neighbours v1->w1, v2->w2, v3->w4 (not a hom., w2w4 not in E(H));
%           best homomorphism v_i -> w_i.
%   H -> G: nearest neighbours send w3, w4 -> v3 (not a hom., w3w4 in E(H));
%           best homomorphism w1->v1, w2->v2, w3->v3, w4->v2.
% ---------------------------------------------------------------------------
\newcommand{\attrspace}{%
  \filldraw[fill = black!4, draw = black!45, line width = 0.4pt]
    (1.0,-0.12) rectangle (6.0,2.0);
  \node[font = \Large, anchor = north east] at (6.0,2.0) {$\mathcal{X}$};
}
\newcommand{\attrpoints}[1][]{%
  \node[gpt = {setA}{black!70}]    (x1) at (1.5,1.20) {};
  \node[gpt = {setA}{black!70}]   (x2) at (4.8,1.15) {$x_2$};
  \node[gpt = {setA}{black!70}]  (x3) at (2.6,0.45) {$x_3$};
  \node[hpt = {setB}{black!70}]    (y1) at (2.0,1.30) {};
  \node[hpt = {setB}{black!70}]   (y2) at (5.30,0.78) {};
  \node[hpt = {setB}{black!70}]  (y3) at (3.7,0.30) {$y_3$};
  \node[hpt = {setB}{black!70}] (y4) at (3.20,1.0) {$y_4$};
}

% Graph H below X (drawn first), graph G above X (drawn last)
\newcommand{\graphH}{%
  \node[hv = setB] (w1) at (2.0,-0.60) {};
  \node[hv = setB] (w2) at (5.30,-0.55) {};
  \node[hv = setB] (w3) at (3.7,-0.82) {};
  \node[hv = setB] (w4) at (3.20,-1.25) {};
  \draw[stem] (w1) -- (y1);
  \draw[stem] (w2) -- (y2);
  \draw[stem] (w3) -- (y3);
  \draw[stem] (w4) -- (y4);
  \draw[gedge] (w1) -- (w2) (w1) -- (w4) (w2) -- (w3) (w3) -- (w4) (w1) -- (w3);
}

\newcommand{\graphG}{%
  \node[gv = setA] (v1) at (1.5,2.80) {};
  \node[gv = setA] (v2) at (4.87,2.9) {};
  \node[gv = setA] (v3) at (2.6,2.32) {};
  \draw[stem] (v1) -- (x1);
  \draw[stem] (v2) -- (x2);
  \draw[stem] (v3) -- (x3);
  \draw[gedge] (v1) -- (v2) (v1) -- (v3) (v2) -- (v3);
}

  % =========================================================================
  % Top row (a): Hausdorff -- every point picks its own nearest neighbour
  % =========================================================================
  % row box + d_H tag (drawn first so it stays in the background)
  \draw[rowbox = colH] (\rowL,-1.15) rectangle (\rowR,2.80);
  \node[rowtag = colH, font=\large] at (\rowL,0.675)
    {(a)\enspace$\mathrm{d}_{\mathrm{H}}$};

  % = max ( panel , panel )   (vertical centre of the content: y = 0.675)
  \node[eqlab, font=\large] at (\parenL-0.03,0.675) {$=\max$};
  \openparen {\parenL}{0.675}{3.5cm}
  \closeparen{\parenR}{0.675}{3.5cm}
  % comma halfway between the two X boxes
  \node[bigcomma] at ({(\panelL+\panelR+8.05)/2},0.60) {$,$};

  % G -> H
  \begin{scope}[xshift = \panelL cm, scale = 1.15]
    \attrspace
    \attrpoints
    \draw[match]  (x1) -- (y1);
    \draw[match] (x2) -- (y2);
    \draw[best] (x3) -- (y4);
    \node[font=\large] at (3.55,-0.55)
      {$\sup\limits_{v \in V} \mathop{\inf\vphantom{\sup}}\limits_{w \in W}\mathrm{d}_{\mathcal{X}}\bigl(f_G(v),f_H(w) \bigr)$};
  \end{scope}
  % H -> G
  \begin{scope}[xshift = \panelR cm, scale = 1.15]
    \attrspace
    \attrpoints
    \draw[match]    (y1) -- (x1);
    \draw[match]   (y2) -- (x2);
    \draw[best]   (y3) -- (x3);
    \draw[match] (y4) -- (x3);
    \node[font=\large] at (3.55,-0.55)
      {$\sup\limits_{w\in W} \mathop{\inf\vphantom{\sup}}\limits_{v\in V}\mathrm{d}_{\mathcal{X}}\bigl(f_G(v),f_H(w) \bigr)$};
  \end{scope}

  % =========================================================================
  % Divider between the two rows
  % =========================================================================
  \draw[divider] (\rowL,-1.30) -- (\rowR,-1.30);

  % =========================================================================
  % Bottom row (b): the same comparisons, restricted to homomorphisms
  % =========================================================================
  \begin{scope}[yshift = -5.6cm]
    % row box + d_HD tag
    \draw[rowbox = colHD] (\rowL,-3.0) rectangle (\rowR,4.15);
    \node[rowtag = colHD, font=\large] at (\rowL,0.525)
      {(b)\enspace$\mathrm{d}_{\mathrm{HD}}$};

    % = max ( panel , panel )   (vertical centre of the content: y = 0.525)
    \node[eqlab, font=\large] at (\parenL-0.03,0.525) {$=\max$};
    \openparen {\parenL}{0.525}{6.35cm}
    \closeparen{\parenR}{0.525}{6.35cm}
    % comma halfway between the two X boxes
    \node[bigcomma] at ({(\panelL+\panelR+8.05)/2},0.45) {$,$};

    % G band (above X) and H band (below X), spanning both panels
    \fill[band = red!8]  (\bandL,2.6)  rectangle (\bandR,3.9);
    \fill[band = blue!8] (\bandL,-0.4) rectangle (\bandR,-1.80);

    % G / H labels + homomorphism direction on the outer side of each panel
    % left panel:  phi : G -> H  (arrow points down, left of the left X box)
    \node[circle, fill=colHD!40, bandlab = black, font=\large]  (GL) at (\panelL+0.8,3.11)  {$G$};
    \node[circle, fill=colHD!40, bandlab = black, font=\large] (HL) at (\panelL+0.8,-1.10) {$H$};
    \draw[homarrow] (GL) -- (HL) node[homlab, midway, left] {$\varphi$};
    % right panel: psi : H -> G  (arrow points up, right of the right X box)
    \node[circle, fill=colHD!40, bandlab = black, font=\large]  (GR) at (\panelR+7.75,3.11)  {$G$};
    \node[circle, fill=colHD!40, bandlab = black, font=\large] (HR) at (\panelR+7.75,-1.10) {$H$};
    \draw[homarrow] (HR) -- (GR) node[homlab, midway, right] {$\psi$};

    % G -> H
    \begin{scope}[xshift = \panelL cm, scale = 1.15]
        \attrspace
        \attrpoints
        \graphH
        \node[lab, left]  at (w1) {$\varphi(v_1)$};
        \node[lab, right] at (w2) {$\varphi(v_2)$};
        \node[lab, below right] at (w3) {$\varphi(v_3)$};
        \node[lab, below right] at (w4) {$w_4$};

        \graphG
        \node[lab, above] at (v1) {$v_1$};
        \node[lab, above right] at (v2) {$v_2$};
        \node[lab, above]  at (v3) {$v_3$};
        \draw[match]  (x1) -- (y1);
        \draw[match] (x2) -- (y2);
        \draw[best] (x3) -- (y3);
        \node[font=\large] at (3.55,-2.1) {$\inf\limits_{\varphi \in \mathfrak{G}} \max\limits_{v \in V} \mathrm{d}_{\mathcal{X}}\bigl( f_G(v), f_H(\varphi(v)) \bigr)$};
    \end{scope}
    % H -> G
    \begin{scope}[xshift = \panelR cm, scale = 1.15]
        \attrspace
        \attrpoints
        \graphH
        \node[lab, left]  at (w1) {$w_1$};
        \node[lab, right] at (w2) {$w_2$};
        \node[lab, below right] at (w3) {$w_3$};
        \node[lab, below right] at (w4) {$w_4$};

        \graphG
        \node[lab, above] at (v1) {$\psi(w_1)$};
        \node[lab, above right] at (v2) {$\psi(w_2)=\psi(w_4)$};
        \node[lab, above=1.5pt]  at (v3) {$\psi(w_3)$};
        \draw[match]    (y1) -- (x1);
        \draw[match]   (y2) -- (x2);
        \draw[match]  (y3) -- (x3);
        \draw[best]  (y4) -- (x2);
        \node[font=\large] at (3.55,-2.1) {$\inf\limits_{\psi \in \mathfrak{H}} \max\limits_{w \in W} \mathrm{d}_{\mathcal{X}}\bigl( f_H(w), f_G(\psi(w)) \bigr)$};
    \end{scope}
  \end{scope}

\end{tikzpicture}

%% file: biblio.bib
@inproceedings{ballester_expressivity_2024,
	title = {On the {Expressivity} of {Persistent} {Homology} in {Graph} {Learning}},
	language = {en},
	author = {Ballester, Rubén and Rieck, Bastian},
    booktitle     = {Proceedings of the Third Learning on Graphs Conference},
    publisher     = {PMLR},
    series        = {Proceedings of Machine Learning Research},
    volume        = 269,
    pages         = {42:1--42:31},
    year          = {2025},
    editor        = {Wolf, Guy and Krishnaswamy, Smita},
}

@inproceedings{boker_complexity_2019,
	title = {The Complexity of Homomorphism Indistinguishability},
	volume = 138,
    booktitle     = {44th International Symposium on Mathematical Foundations of Computer Science~(MFCS)},
    publisher     = {Schloss Dagstuhl -- Leibniz-Zentrum f{\"u}r Informatik},
    editor        = {Rossmanith, Peter and Heggernes, Pinar and Katoen, Joost-Pieter},
    series        = {Leibniz International Proceedings in Informatics (LIPIcs)},
	author = {Böker, Jan and Chen, Yijia and Grohe, Martin and Rattan, Gaurav},
	collaborator = {Rossmanith, Peter and Heggernes, Pinar and Katoen, Joost-Pieter},
	year = {2019},
	pages = {54:1--54:13},
}

@article{daaz_counting_2002,
	title = {Counting \mbox{$H$-colorings} of partial $k$-trees},
	language = {en},
	journal = {Theoretical Computer Science},
    author        = {Josep D{\'i}az and Maria Serna and Dimitrios M. Thilikos},
	year = 2002,
}

@inproceedings{curticapean_homomorphisms_2017,
	title = {Homomorphisms Are a Good Basis for Counting Small Subgraphs},
	language = {en},
	urldate = {2025-02-13},
	booktitle = {Proceedings of the 49th {Annual} {ACM} {SIGACT} {Symposium} on {Theory} of {Computing}},
	author = {Curticapean, Radu and Dell, Holger and Marx, Dániel},
	year = {2017},
	pages = {210--223},
}

@article{bodlaender_combinatorial_2007,
	title = {Combinatorial Optimization on Graphs of Bounded Treewidth},
	volume = {51},
	language = {en},
	number = {3},
	journal = {The Computer Journal},
	author = {Bodlaender, H. L. and Koster, A. M. C. A.},
	year = {2007},
	pages = {255--269},
}

@incollection{hutchison_fine_2013,
	address = {Cham, Switzerland},
	title = {The Fine Details of Fast Dynamic Programming over Tree Decompositions},
	volume = {8246},
	language = {en},
	booktitle = {Parameterized and Exact Computation},
	publisher = {Springer},
	author = {Bodlaender, Hans L. and Bonsma, Paul and Lokshtanov, Daniel},
	editor = {Hutchison, David and Kanade, Takeo and Kittler, Josef and Kleinberg, Jon M. and Mattern, Friedemann and Mitchell, John C. and Naor, Moni and Nierstrasz, Oscar and Pandu Rangan, C. and Steffen, Bernhard and Sudan, Madhu and Terzopoulos, Demetri and Tygar, Doug and Vardi, Moshe Y. and Weikum, Gerhard and Gutin, Gregory and Szeider, Stefan},
	year = {2013},
	series = {Lecture Notes in Computer Science},
	pages = {41--53},
}

@inproceedings{wang_empirical_2024,
  title         = {An Empirical Study of Realized {GNN} Expressiveness},
  author        = {Wang, Yanbo and Zhang, Muhan},
  year          = 2024,
  booktitle     = {Proceedings of the 41st International Conference on Machine Learning},
  publisher     = {PMLR},
  series        = {Proceedings of Machine Learning Research},
  volume        = 235,
  pages         = {52134--52155},
  editor        = {Salakhutdinov, Ruslan and Kolter, Zico and Heller, Katherine and Weller, Adrian and Oliver, Nuria and Scarlett, Jonathan and Berkenkamp, Felix}
}

@inproceedings{bao_homomorphism_2025,
	title = {Homomorphism Counts as Structural Encodings for Graph Learning},
	language = {en},
    booktitle     = {International Conference on Learning Representations},
	author = {Bao, Linus and Jin, Emily and Bronstein, Michael M. and Ceylan, Ismail Ilkan and Lanzinger, Matthias},
	year = {2025},
}

@inproceedings{jin_homomorphism_2024,
  title         = {Homomorphism Counts for Graph Neural Networks: All About That Basis},
  author        = {Jin, Emily and Bronstein, Michael M. and Ceylan, Ismail Ilkan and Lanzinger, Matthias},
  year          = 2024,
  booktitle     = {Proceedings of the 41st International Conference on Machine Learning},
  publisher     = {PMLR},
  series        = {Proceedings of Machine Learning Research},
  volume        = 235,
  pages         = {22075--22098},
  editor        = {Salakhutdinov, Ruslan and Kolter, Zico and Heller, Katherine and Weller, Adrian and Oliver, Nuria and Scarlett, Jonathan and Berkenkamp, Felix}
}

@book{lovasz_large_2012,
	address = {Providence, RI},
	series = {Colloquium {Publications}},
	title = {Large {Networks} and {Graph} {Limits}},
	volume = {60},
	language = {en},
	urldate = {2025-04-28},
	publisher = {American Mathematical Society},
	author = {Lov{\'a}sz, László},
	year = {2012},
}

@inproceedings{welke2023expectation,
  title         = {Expectation-Complete Graph Representations with Homomorphisms},
  author        = {Welke, Pascal and Thiessen, Maximilian and Jogl, Fabian and G\"{a}rtner, Thomas},
  year          = 2023,
  booktitle     = {Proceedings of the 40th International Conference on Machine Learning},
  publisher     = {PMLR},
  series        = {Proceedings of Machine Learning Research},
  volume        = 202,
  pages         = {36910--36925},
  editor        = {Krause, Andreas and Brunskill, Emma and Cho, Kyunghyun and Engelhardt, Barbara and Sabato, Sivan and Scarlett, Jonathan}
}

@article{wolf_structural_2023,
	title = {Structural {Node} {Embeddings} with {Homomorphism} {Counts}},
	language = {en},
	publisher = {arXiv},
	author = {Wolf, Hinrikus and Oeljeklaus, Luca and Kühner, Pascal and Grohe, Martin},
	year = {2023},
    journal={arXiv preprint arXiv:2308.15283},
}

@article{morris_weisfeiler_2019,
	title = {Weisfeiler and {Leman} Go Neural: Higher-Order Graph Neural Networks},
	volume = {33},
	doi = {10.1609/aaai.v33i01.33014602},
	language = {en},
	number = {01},
	urldate = {2025-06-02},
	journal = {Proceedings of the AAAI Conference on Artificial Intelligence},
	author = {Morris, Christopher and Ritzert, Martin and Fey, Matthias and Hamilton, William L. and Lenssen, Jan Eric and Rattan, Gaurav and Grohe, Martin},
	year = {2019},
	pages = {4602--4609},
}

@inproceedings{kanatsoulis2025learning,
  title         = {Learning Efficient Positional Encodings with Graph Neural Networks},
  author        = {Charilaos Kanatsoulis and Evelyn Choi and Stefanie Jegelka and Jure Leskovec and Alejandro Ribeiro},
  year          = 2025,
  booktitle     = {International Conference on Learning Representations},
}

@article{dwivedi2023benchmarking,
  title={Benchmarking graph neural networks},
  author={Dwivedi, Vijay Prakash and Joshi, Chaitanya K and Luu, Anh Tuan and Laurent, Thomas and Bengio, Yoshua and Bresson, Xavier},
  journal={Journal of Machine Learning Research},
  volume={24},
  number={43},
  pages={1--48},
  year={2023}
}

@inproceedings{dwivedi2021graph,
  title={Graph neural networks with learnable structural and positional representations},
  author={Dwivedi, Vijay Prakash and Luu, Anh Tuan and Laurent, Thomas and Bengio, Yoshua and Bresson, Xavier},
  journal={arXiv preprint arXiv:2110.07875},
  year=2022,
  booktitle     = {International Conference on Learning Representations},
}

@inproceedings{nguyen2020graph,
  title         = {Graph Homomorphism Convolution},
  author        = {Nguyen, Hoang and Maehara, Takanori},
  year          = 2020,
  booktitle     = {Proceedings of the 37th International Conference on Machine Learning},
  publisher     = {PMLR},
  series        = {Proceedings of Machine Learning Research},
  volume        = 119,
  pages         = {7306--7316},
  editor        = {Daum{\'e}~III, Hal and Singh, Aarti}
}

@inproceedings{xu2018powerful,
  title         = {How Powerful are Graph Neural Networks?},
  author        = {Keyulu Xu and Weihua Hu and Jure Leskovec and Stefanie Jegelka},
  year          = 2019,
  booktitle     = {International Conference on Learning Representations},
}

@Article{lovasz1967operations,
author={Lov{\'a}sz, L.},
title={Operations with structures},
journal={Acta Mathematica Academiae Scientiarum Hungarica},
year={1967},
volume={18},
number={3},
pages={321--328},
}

@book{hell2004graphs,
  title= {Graphs and Homomorphisms},
  author={Hell, Pavol and Nesetril, Jaroslav},
  year={2004},
  publisher={Oxford University Press},
}

@inproceedings{joshi2023expressive,
  title         = {On the Expressive Power of Geometric Graph Neural Networks},
  author        = {Joshi, Chaitanya K. and Bodnar, Cristian and Mathis, Simon V and Cohen, Taco and Lio, Pietro},
  year          = 2023,
  booktitle     = {Proceedings of the 40th International Conference on Machine Learning},
  publisher     = {PMLR},
  series        = {Proceedings of Machine Learning Research},
  volume        = 202,
  pages         = {15330--15355},
  editor        = {Krause, Andreas and Brunskill, Emma and Cho, Kyunghyun and Engelhardt, Barbara and Sabato, Sivan and Scarlett, Jonathan},
}

@article{morris2023weisfeiler,
  title         = {{W}eisfeiler and {L}eman go Machine Learning: The Story so far},
  author        = {Christopher Morris and Yaron Lipman and Haggai Maron and Bastian Rieck and Nils M. Kriege and Martin Grohe and Matthias Fey and Karsten Borgwardt},
  year          = 2023,
  journal       = {Journal of Machine Learning Research},
  volume        = 24,
  number        = 333,
  pages         = {1--59},
}

@inproceedings{bechlerspeicher25a,
  title         = {Position: Graph Learning Will Lose Relevance Due To Poor Benchmarks},
  author        = {Bechler-Speicher, Maya and Finkelshtein, Ben and Frasca, Fabrizio and M\"{u}ller, Luis and T\"{o}nshoff, Jan and Siraudin, Antoine and Zaverkin, Viktor and Bronstein, Michael M. and Niepert, Mathias and Perozzi, Bryan and Galkin, Mikhail and Morris, Christopher},
  year          = 2025,
  booktitle     = {Proceedings of the 42nd International Conference on Machine Learning},
  publisher     = {PMLR},
  series        = {Proceedings of Machine Learning Research},
  volume        = 267,
  pages         = {81067--81089},
  editor        = {Singh, Aarti and Fazel, Maryam and Hsu, Daniel and Lacoste-Julien, Simon and Berkenkamp, Felix and Maharaj, Tegan and Wagstaff, Kiri and Zhu, Jerry},
}

@inproceedings{coupette25a,
  title         = {No Metric to Rule Them All: Toward Principled Evaluations of Graph-Learning Datasets},
  author        = {Coupette, Corinna and Wayland, Jeremy and Simons, Emily and Rieck, Bastian},
  year          = 2025,
  booktitle     = {Proceedings of the 42nd International Conference on Machine Learning},
  publisher     = {PMLR},
  series        = {Proceedings of Machine Learning Research},
  volume        = 267,
  pages         = {11405--11434},
  editor        = {Singh, Aarti and Fazel, Maryam and Hsu, Daniel and Lacoste-Julien, Simon and Berkenkamp, Felix and Maharaj, Tegan and Wagstaff, Kiri and Zhu, Jerry},
}

@article{ivanov2019understanding,
  title={Understanding isomorphism bias in graph data sets},
  author={Ivanov, Sergei and Sviridov, Sergei and Burnaev, Evgeny},
  journal={arXiv preprint arXiv:1910.12091},
  year={2019}
}

@inproceedings{babai1983canonical,
  title={Canonical labeling of graphs},
  author={Babai, L{\'a}szl{\'o} and Luks, Eugene M},
  booktitle={Proceedings of the 15th Annual ACM Symposium on Theory of Computing},
  pages={171--183},
  year={1983}
}

@article{vasileiou2026graph,
  title={Graph Learning Should Move Beyond Restrictive Views of Spectral and Message-Passing {GNNs}},
  author={Vasileiou, Antonis and Cervino, Juan and Frossard, Pascal and Kanatsoulis, Charilaos I and Morris, Christopher and Schaub, Michael T and Vandergheynst, Pierre and Wang, Zhiyang and Wolf, Guy and Levie, Ron},
  journal={arXiv preprint arXiv:2602.10031},
  year={2026}
}

@article{hu2020open,
  title={Open graph benchmark: Datasets for machine learning on graphs},
  author={Hu, Weihua and Fey, Matthias and Zitnik, Marinka and Dong, Yuxiao and Ren, Hongyu and Liu, Bowen and Catasta, Michele and Leskovec, Jure},
  journal={Advances in Neural Information Processing Systems},
  volume={33},
  pages={22118--22133},
  year={2020}
}

@article{isufi2024graph,
  title={Graph filters for signal processing and machine learning on graphs},
  author={Isufi, Elvin and Gama, Fernando and Shuman, David I and Segarra, Santiago},
  journal={IEEE Transactions on Signal Processing},
  volume={72},
  pages={4745--4781},
  year={2024},
  publisher={IEEE}
}

@inproceedings{guo2025position,
  title         = {Position: Spectral {GNN}s Rely Less on Graph {F}ourier Basis than Conceived},
  author        = {Guo, Yuhe and Tang, Huayi and Ma, Jiahong and Xu, Hongteng and Wei, Zhewei},
  year          = 2025,
  booktitle     = {Proceedings of the 42nd International Conference on Machine Learning},
  publisher     = {PMLR},
  series        = {Proceedings of Machine Learning Research},
  volume        = 267,
  pages         = {81471--81493},
  editor        = {Singh, Aarti and Fazel, Maryam and Hsu, Daniel and Lacoste-Julien, Simon and Berkenkamp, Felix and Maharaj, Tegan and Wagstaff, Kiri and Zhu, Jerry},
}

@article{vayer2020fused,
  title={Fused {G}romov-{W}asserstein distance for structured objects},
  author={Vayer, Titouan and Chapel, Laetitia and Flamary, R{\'e}mi and Tavenard, Romain and Courty, Nicolas},
  journal={Algorithms},
  volume={13},
  number={9},
  pages={212},
  year={2020},
  publisher={MDPI}
}

@article{kempton2024isospectral,
  title={Isospectral reductions and quantum walks on graphs},
  author={Kempton, Mark and Tolbert, John},
  journal={Algebraic Combinatorics},
  volume={7},
  number={1},
  pages={225--243},
  year={2024}
}

@article{morris2020tudataset,
  title={Tudataset: A collection of benchmark datasets for learning with graphs},
  author={Morris, Christopher and Kriege, Nils M and Bause, Franka and Kersting, Kristian and Mutzel, Petra and Neumann, Marion},
  journal={arXiv preprint arXiv:2007.08663},
  year={2020}
}

@book{kloks1994treewidth,
  title={Treewidth: computations and approximations},
  author={Kloks, Ton},
  year={1994},
  publisher={Springer}
}

@misc{hagberg2020networkx,
  title={Networkx: Network analysis with {P}ython},
  author={Hagberg, Aric and Conway, Drew},
  url={https://networkx. github. io},
  volume={1031},
  year={2020}
}

@article{arvind2020weisfeiler,
  title={On {W}eisfeiler-{L}eman invariance: Subgraph counts and related graph properties},
  author={Arvind, Vikraman and Fuhlbr{\"u}ck, Frank and K{\"o}bler, Johannes and Verbitsky, Oleg},
  journal={Journal of Computer and System Sciences},
  volume={113},
  pages={42--59},
  year={2020},
  publisher={Elsevier}
}
